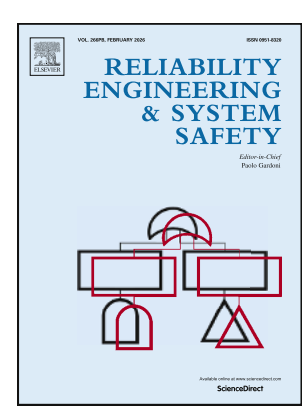

# Towards reliable multimodal disaster severity assessment through preference optimization and explainable vision-language reasoning

Yuanjun Zhang [a,1], Fuzel Ahamed Shaik [a,1,*], Suvojit Acharjee [b], Fahad Khalid [a], Mourad Oussalah [a]

[a] Centre for Machine Vision & Signal Processing, University of Oulu, Pentti Kaiteran katu 1, PO Box 8000, Oulu, 90570, Finland
[b] Department of CSE and Institute of Engineering and Management, 120 SDF Building, Saltlake Electronics Complex, Kolkata, 700091, India

ABSTRACT

Reliable disaster damage assessment requires models that provide both accurate predictions and transparent explanations. However, existing multimodal approaches are limited by scarce annotated data and insufficient evaluation of reasoning quality. This study proposes a two-stage training framework that integrates Supervised Fine-Tuning (SFT) and Direct Preference Optimization (DPO) within a unified data construction pipeline. From a single Human-in-the-Loop (HITL) annotation workflow, two complementary datasets are derived, namely ReasoningSet, which contains validated rationales for SFT, and PreferenceSet, which comprises paired rationales for DPO-based alignment. The framework evaluates both classification performance and explanation quality using automatic metrics, model-based scoring, and human ranking. Experimental results show that SFT improves accuracy from 73.64% to 78.29% and increases Macro-F1 by 29% compared to the baseline, while explanation quality improves by approximately 25%. Subsequent DPO alignment further enhances interpretability on the PreferenceSet. Cross-model validation on InternVL-3-8B and LLaVA-1.5-7B demonstrates the robustness and generalizability of the approach. The proposed framework improves detection of underrepresented mild damage cases, reduces high-risk misclassifications, and strengthens alignment between model reasoning and human judgment. Overall, it provides a reproducible pathway to develop reliable multimodal systems that deliver auditable, actionable disaster insights for emergency management.

## 1. Introduction

The increasing dominance of digital presence among citizens has motivated emergency service providers, humanitarian organizations, and policymakers to rely on social networks as a crucial medium for tracking real-time events [1]. Social media enables the rapid dissemination of information to broad and diverse audiences [2], a feature that becomes particularly vital in the context of disasters, where timeliness directly influences the effectiveness of response and mitigation efforts. These platforms not only allow authorities to distribute instructions but also to receive situational reports, requests for assistance, and localized updates from affected individuals, thus improving situational awareness, coordination, and citizen participation during crises. The vast and unregulated nature of user-generated content presents serious challenges, such as posts being frequently noisy, unverified, and prone to misinformation, which can exacerbate confusion or hinder decision-making in high-stakes environments, an emerging reliability concern as disinformation increasingly affects critical infrastructure and decision chains [3]. Addressing these issues has therefore emerged as a critical research priority in crisis informatics and emergency communication, underscoring the need for analytical frameworks that can balance the immediacy of social media-based evidence with the imperatives of accuracy, trust, and reliability. In the reliability and safety community, these needs are typically framed as risk-informed decision support for emergency management, where models must be auditable and operationally useful under time pressure [4].

In practical emergency response settings, damage assessments derived from social media can influence the prioritization of rescue operations, allocation of medical resources, and deployment of inspection teams. Inaccurate or non-transparent model outputs may lead to misallocation, delayed intervention, or erosion of institutional trust. As social sensing increasingly supplements traditional monitoring systems, ensuring that automated assessments are both accurate and explainable becomes not only a technical requirement but a societal necessity. The reliability of such systems directly affects public safety, infrastructure resilience, and humanitarian effectiveness.

* Corresponding author.
*E-mail address:* fuzel.shaik@oulu.fi (F.A. Shaik).
[1] These authors have contributed equally to this work.

A growing body of research in crisis informatics has highlighted both the potential and limitations of using social media data, particularly from Platform X (formerly Twitter), for disaster management [6,7]. Due to its immediacy, brevity, and global reach, Platform X has been extensively studied for early situational awareness in crises such as earthquakes, floods, wildfires, and hurricanes, with evidence that its data can support event detection, spatiotemporal tracking, and disaster impact assessment from textual and multimedia content [8,9]. However, operational adoption remains hindered by quality, reliability, and representativeness issues. These challenges are often amplified when incorporating images but also raises concerns about authenticity and annotation reliability, despite encouraging progress in multimodal analysis [11]. For safety analysis and emergency management, this multimodal evidence is increasingly fed into some model-based risk assessment and operational twins for planning and response [12]. In practice, publicly available datasets such as CrisisMMD [13] and more recent multilingual resources including BanglaCalamityMMD [14] have enabled multimodal frameworks for damage classification [15,16]. However, most such models operate as black boxes, delivering only categorical labels (e.g., *severe*, *mild*, or *little or no damage*) without revealing the underlying rationale for the result. This lack of transparency often limits real-world deployment. Respondents must be able to verify why a prediction was made to build trust, support error analysis, and integrate domain expertise [17,18]. Consequently, there is a pressing need for systems that can both predict and explain, answering not only *what* level of damage is present but also *why* the available evidence supports that assessment, thereby improving the reliability of existing decision support systems in emergency management. Such interventions can be framed as risk-informed decision support for emergency management [5,12]. In this context, this paper is guided by three research questions:

- **RQ1:** How can multimodal crisis datasets be extended with human-validated rationale to support the development of explainable models for disaster damage assessment?
- **RQ2:** What training and alignment strategies enable vision language models (VLMs) to jointly predict disaster severity and provide human-aligned explanations that improve trust, reliability, and usability?
- **RQ3:** How can evaluation protocols be designed to systematically measure both predictive performance and explanation quality, ensuring that outputs are actionable for practitioners in real-world crisis response settings?

Methodologically, this study adopts a mixed-method strategy that integrates large-scale computational modeling with structured human evaluation and preference-based alignment. Automated generation using state-of-the-art vision-language models is complemented by a multi-stage human-in-the-loop validation protocol involving trained contributors and expert reviewers. This design enables both quantitative performance assessment and qualitative explanation auditing, bridging machine learning development with human-centered reliability requirements. By explicitly combining algorithmic training, human validation, and preference optimization, the framework addresses the limitations of purely automated pipelines while ensuring operational interpretability in safety-critical decision contexts. Despite significant advances in multimodal disaster damage classification, three structural limitations remain. First, existing crisis datasets such as CrisisLexT26 and CrisisMMD provide severity or informativeness labels without accompanying rationales, leaving the underlying decision logic implicit and inaccessible for downstream auditing or expert verification [10,13,15]. Second, most multimodal disaster assessment models prioritize classification performance, employing late- or early-fusion architectures that yield accurate predictions but offer limited insight into the causal contribution of textual or visual evidence [11,16]. As noted by Basit et al. [17], such systems often generate post-hoc explanations that are weakly grounded and insufficient for operational trust. Third, while explainable AI has gained attention, only few studies in crisis informatics adopt structured evaluation protocols for explanation quality, relying instead on informal visualization or anecdotal inspection. This limitation has been widely recognized in the broader interpretability literature, which emphasizes the need for task-aligned, human-centered, and auditable evaluation methods [5,12,19]. The absence of human-validated rationales prevents systematic alignment between machine outputs and practitioner expectations. This gap motivates the development of a reasoning-augmented dataset and alignment framework that explicitly integrates explanation generation, human preference modeling, and auditability.

To address this gap, we introduce CrisisMMD-R, the first reasoning dataset for crisis damage assessment, which extends the original CrisisMMD corpus by adding human-validated explanations that link textual and visual cues to severity labels. Unlike prior datasets that provide only categorical annotations, CrisisMMD-R offers paired rationales showing why a given text-image post is mapped to a particular level of damage. By extension, the use of social media for damage assessment can be viewed as an emerging sensing layer within the resilience engineering and reliability framework of socio-technical critical systems [20]. We adopt a mixed-method approach where the synthetic explanations are first generated automatically using the state-of-the-art vision language model Qwen-VL-Max, and then iteratively improved through a multi-stage human-in-the-loop (HITL) protocol where trained contributors score and provide feedback on drafts, while professional reviewers edit and refine them according to a structured evaluation framework. The reviewers validate each case provided by the contributors, producing aligned pairs of preferred and rejected rationales that can be used to further refine the model using techniques such as Direct Preference Optimization [21,22]. This design allows the dataset to serve both as a training resource for explainable model development and as a benchmark for systematically evaluating explanation quality.

Building on this resource, we fine-tune open-source VLMs to jointly predict and explain the severity of the disaster category. Specifically, our main implementation uses Qwen2.5-VL-7B [23], enhanced with supervised fine-tuning (SFT) and preference-based alignment to ensure reliability and trustworthiness. To further prove the generalizability of our framework, we also conduct comparative experiments on representative open-source models, InternVL3-8B [24], and LLaVA1.5-7B [25]. By framing the task as a hybrid of classification and explanation, we enable models not only to identify whether damage is severe, mild, or little or no damage, but also to articulate the evidence that supports their judgment. This unified formulation supports richer evaluation protocols that combine traditional classification metrics with assessments of explanation quality, including automated measures such as BLEU and ROUGE, external judgments from advanced models, and human ranking to ensure consistency and reduce bias, aligning scenario-based emergency decision-making frameworks used in reliability engineering [19]. Together, these contributions advance crisis informatics from predictive classification toward auditable, explanation-aware decision support systems capable of integration into reliability-sensitive emergency management workflows. Overall, the primary contributions of this study can be summarized as follows.

- We introduce CrisisMMD-R, the first dataset of multimodal crisis posts with human-validated, fine-grained explanations and aligned Preferred and Rejected pairs suitable for alignment research.
- We present a scalable human-in-the-loop annotation protocol with an explicit evaluation framework for contributors and reviewers, incorporating a transparent audit trail and continuous quality improvement mechanisms, specifically tailored to disaster management handling.
- We develop an explainable vision-language model based on the state-of-the-art Qwen2.5-VL-7B, fine-tuned with supervised and preference-based methods to generate both labels and explanations.

- We propose a comprehensive evaluation framework that jointly assesses classification performance and explanation quality, advancing the development of explainable multimodal models for disaster response management.

The remainder of this paper is structured as follows. Section 2 reviews the relevant literature on the current workflow practices, popular damage assessment techniques, VLMs for disaster response, and human enablers to increase trust. Section 3 provides a detailed explanation of the adopted methodology to address our research questions. Section 4 illustrates the experimental results and discusses the experimental performance. Finally, Section 5 summarizes the key findings and draws on the previous works.

## 2. Related work

The following section situates our study within the broader landscape of crisis informatics, multimodal disaster assessment, and explainable AI. We first grounded the discussion in foundational research on Disaster Risk Reduction (DRR) and practitioner-centered emergency workflows, highlighting operational constraints such as credibility, accountability, and system resilience. We then review existing crisis datasets and Vision Language Models (VLMs) approaches, emphasizing both their contributions and their limitations in safety-critical humanitarian contexts. Building on this foundation, we examine advances in explainable multimodal reasoning and HITL alignment, including preference optimization techniques that enable models to generate transparent, auditable justifications. Across these strands of literature, we identify a consistent gap, the absence of multimodal disaster datasets with validated human rationales that support both interpretability and practitioner trust. This section clarifies how CrisisMMD-R emerges at the intersection of crisis informatics, explainable AI, and alignment research, positioning our contribution as a step toward accountable, reasoning-aware decision support in emergency response systems.

### *2.1. Foundations of crisis informatics and practitioner workflows*

DRR frameworks such as the Sendai Framework emphasize risk-informed decision-making across preparedness, response, and recovery phases and call for information systems that are timely, reliable, and accountable [26]. Crisis informatics operationalizes these goals by studying how people, organizations, and technologies interact in emergencies and translating those insights into tools that improve situational awareness and resource allocation. A substantial body of work shows that social media can serve as citizen sensors during crises, enabling early event detection, needs assessment, and rapid dissemination of warnings [27–29]. Volunteer and Technical Communities (V&TCs) have institutionalized these practices via digital humanitarian infrastructures (e.g., Ushahidi deployments for crowdsourced mapping and triage), which bridge public reports with operational workflows in emergency operations centers [30–32]. These infrastructures complement remote sensing and official reporting, but also introduce data quality and governance challenges at scale. Recent studies also integrate System-Theoretic Accident Model and Processes (STAMP) with dynamic Bayesian networks to assess the resilience of emergency response systems, promoting auditability and system-level controls beyond component failure views [33]. This explicitly addresses emergency response systems, socio-technical interactions (human/organization/technology), resilience assessment, and dynamic modeling.

Two persistent barriers limit operational uptake of social media analytics , which are credibility and accountability. Credibility is rooted in rumors and partially verified claims that proliferate during fast-moving events and false content that can diffuse faster and further than verified reports [34,35]. Accountability requires emergency managers to use assessments that are not only accurate but also auditable, so that model outputs can be inspected, trusted, and incorporated into existing decision pipelines and after-action reviews [31]. The present work is grounded in these operational realities. Whereas prior crisis datasets typically offer labels without explanations, our approach augments multimodal posts (text and images) with human-validated rationales that explicitly link textual and visual cues to the assigned damage level. This bridges credibility and accountability gaps identified above by (i) making classification decisions inspectable and (ii) aligning outputs with practitioner information needs (e.g., evidence-based justifications that support triage and tasking). The added rationale signals also create new opportunities for alignment and preference optimization, enabling models to better conform to humanitarian standards for trustworthy decision support.

### *2.2. Crisis informatics and damage-assessment datasets*

Early work on mining social media for emergency management focused on text-only datasets such as CrisisLex [36] and CrisisNLP [10], which collected and annotated tweets across a range of natural disasters. These datasets enabled foundational research on information categorization, situational awareness, and humanitarian information needs [28,29]. With the growing recognition of visuals as critical evidence in crises, CrisisMMD was released [13], pairing tweets with images and assigning a three-level damage severity label that has since become a cornerstone for multimodal disaster classification. More recently, efforts have expanded multimodal disaster datasets beyond English-centric contexts. For example, BanglaCalamityMMD [14] introduces a comprehensive benchmark for multimodal disaster identification in the low-resource Bangla language, highlighting the importance of linguistic diversity and inclusivity in crisis informatics research. Such multilingual benchmarks significantly broaden the applicability of multimodal disaster detection systems and emphasize the need for culturally adaptive models. UAV-based imagery has been used for fine-grained post-disaster damage mapping [37], while satellite-based datasets have been developed to capture infrastructure disruptions and large-scale damage patterns [38,39]. More recent work has also explored crowdsourced mapping platforms such as OpenStreetMap for disaster response [32], reflecting the increasing diversity of geospatial data sources. Such heterogeneous sources are also coupled with probabilistic resilience models (e.g., BN/DBN-based formulations) to support reliability-oriented disaster analysis and planning [40]. The absence of explicit rationales limits both the interpretability of models and the ability of practitioners to integrate machine outputs into decision-making. In safety-critical settings, the lack of auditable reasoning further constrains risk-informed emergency scheme selection and post-event accountability [41]. Our work fills this gap by augmenting CrisisMMD with crowd and expert-validated explanations, enabling systematic research on explainable crisis assessment.

### *2.3. Vision-language models for disaster response*

Mainstream VLMs such as ViLBERT [42], VisualBERT [43], CLIP [44], BLIP-2 [45], and Flamingo [46] have demonstrated strong generalization to a wide range of multimodal tasks. Their success has encouraged exploration of zero-shot and few-shot learning for specialized domains [47,48]. However, disaster response introduces unique requirements, such as predictions must be not only accurate but also trustworthy, as decisions can have direct humanitarian consequences. Recent studies underscore this by framing situation awareness as dynamic, risk-informed inference with Bayesian networks tailored to emergency response information-scapes (ERIMap) [49]. Most existing VLMs remain oriented toward label prediction without generating explicit reasoning chains. Instruction-tuned models such as Qwen-VL [50] and LLaVA [51] introduce the ability to verbalize rationales, yet no work has aligned such models specifically for crisis informatics. Prior studies that employ VLMs for disasters often focus on tasks such as remote sensing image classification [52], but do not address multimodal social media data where both text and imagery jointly shape interpretation. We build on Qwen2.5-VL-7B [23] and show that targeted fine-tuning with disaster-specific rationales, combined with preference optimization, markedly improves both classification accuracy and explanation quality in this domain. In addition, we include representative open-source models such as InternVL3-8B [24] and LLaVA1.5-7B [25] for cross-validation.

### 2.4. Explainable multimodal reasoning

The broader field of explainable AI (XAI) has long emphasized the role of rationales in exposing model reasoning to human users. Textual rationales have been introduced for natural language inference through datasets such as e-SNLI [53] and WT5 [54], which encourage models to justify their predictions alongside labels. For vision-language tasks, [55] developed VQA-X, supplying sentence-level explanations for visual question answering, while Do et al. [56] introduced e-SNLI-VE for visual entailment. Other work has combined textual justifications with visual grounding to provide multimodal transparency, exemplified by REx [57]. More recently, research has focused on fine-grained alignment between model outputs and human preferences, using natural language justifications in science QA [58] and multimodal reasoning in healthcare [59]. Despite these advances, no dataset offers rationales for disaster-damage classification, a domain where interpretability is mission-critical due to the high stakes of humanitarian decision-making. Our CrisisMMD-R dataset extends the philosophy of self-rationalization to crisis informatics, pairing each label with a human-validated explanation and an associated quality score. These annotations can be directly exploited by alignment algorithms, allowing models to learn not only what the correct answer is but also why. From a reliability perspective, pairing labels with audited rationales facilitates system-level resilience assessment and structured after-action reviews [40].

### 2.5. Human-in-the-loop alignment and preference optimization

Recent progress in aligning Large Language Models (LLMs) with human expectations has been driven by Reinforcement Learning from Human Feedback (RLHF) [60], as well as its non-RL variant Direct Preference Optimization (DPO) [21]. These methods rely on human judgments to create preference pairs that guide models toward generating more desirable outputs. Several works have begun to adapt these ideas to vision-language models. For example, LLaVA [51] and BLIP-2 [45], though most training signals come from synthetic instructions or curated prompts rather than explicit preference-based rankings. An exception is Illume [61], which demonstrates the value of HITL evaluation for multimodal models. In parallel, parameter-efficient fine-tuning techniques such as LoRA [62] and its variants [63] have enabled scalable adaptation of billion-parameter models in specialized domains, significantly reducing the computational cost of alignment. In the disaster response setting, where training data is scarce and domain expertise is critical, combining expert verification with structured crowd editing provides a cost-effective pathway to produce high-quality preferred and rejected pairs. Our protocol builds on this line of work by generating transparent rationales for crisis data and using them to fine-tune and align VLMs through DPO. This approach allows models to better reflect practitioner needs, balancing predictive performance with interpretability and accountability.

## 3. Methodology

This section presents the complete methodological pipeline that transforms raw multimodal crisis data into an explainable and reliability-oriented vision-language model. Our design emphasizes not only interpretability but also data and reasoning reliability from dataset construction to model alignment. It first details how data resources are constructed and expanded by prompt-based rationale generation, then formalizes the joint classification-with-explanation objective, followed by the base architecture and the two-stage training recipe (LoRA-SFT → DPO). An overview of the entire pipeline is illustrated in Fig. 1.

**Table 1**
Original CrisisMMD damage-level counts.

| | Categories | Number |
|---|---|---|
| Damage Severity | Little or No Damage | 475 |
| | Mild Damage | 839 |
| | Severe Damage | 2212 |
| | Total | 3526 |

### 3.1. Dataset construction

#### 3.1.1. Overview of data construction pipeline

The three green circles in the left panel of Fig. 1 represent the original, clean components of the CrisisMMD dataset, each denoting verified and noise-free inputs. In contrast, the orange nodes labeled “Reasoning” indicate the additional explanatory dimension introduced by our extension. At the beginning of the pipeline, these reasoning entries are heterogeneous. The lighter-to-darker orange gradient symbolizes noisy, machine-generated drafts produced by Qwen-VL-Max under a carefully designed prompting scheme. Since not all model outputs are trustworthy, this initial reasoning pool contains both high-quality and flawed explanations.

These raw drafts are subsequently processed through a HITL pipeline comprising evaluation, correction, and professional review. During this multi-stage curation, explanations that already satisfy the evaluation criteria are directly accepted as clean, validated reasonings (depicted as the fully saturated dark-orange nodes), forming the basis of the ReasoningSet for Supervised Fine-tuning (SFT). Drafts that fail to pass the quality threshold, on the other hand, are sent back for human expert correction. The corrected versions become new dark-orange entries, while their original, uncorrected counterparts remain as the lighter, noisy variants. Each such pair, machine-generated (rejected) versus human-revised (preferred), constitutes a preference pair in the PreferenceSet used for DPO. This design ensures that the ReasoningSet and PreferenceSet are mutually exclusive, which means that the verified reasonings contribute only to supervised training, whereas edited pairs exclusively serve preference alignment. Together, they enable CrisisMMD-R to support both supervised instruction tuning and human-aligned optimization within a unified methodological framework. Subsequent subsections provide implementation details for each stage of the pipeline.

#### 3.1.2. CrisisMMD: Multimodal crisis dataset

The Damage-Severity split of CrisisMMD comprises 3526 tweet-image pairs drawn from seven natural disasters. The class distribution is reproduced verbatim in Table 1. No explanatory text accompanies the labels, making CrisisMMD a natural test-bed for explanation augmentation.

#### 3.1.3. Prompt engineering and draft generation

To elicit label-consistent rationales we designed a chain-of-thought prompt that (i) restates CrisisMMD’s official criteria, (ii) clarifies that tweets are only secondary evidence, and (iii) constrains length to facilitate later fine-tuning. The final prompt is shown in Fig. 2. Feeding every CrisisMMD item to Qwen-VL-Max with this prompt yields 3526 draft explanations, one per tweet-image pair. As illustrated in Fig. 3 in the OpenCompass multimodal leaderboard maintained by VLMEvalKit [64], Qwen-VL-Max ranks near the very top across ten diverse benchmarks, displaying performance that is competitive with the leading proprietary systems while outperforming all other open-weight models in most cases. This favorable accuracy profile, together with its comparatively lower usage fees, makes Qwen-VL-Max the natural choice for producing the initial rationales in our pipeline. The reduced cost is particularly valuable for our study, as it enables us to generate thousands

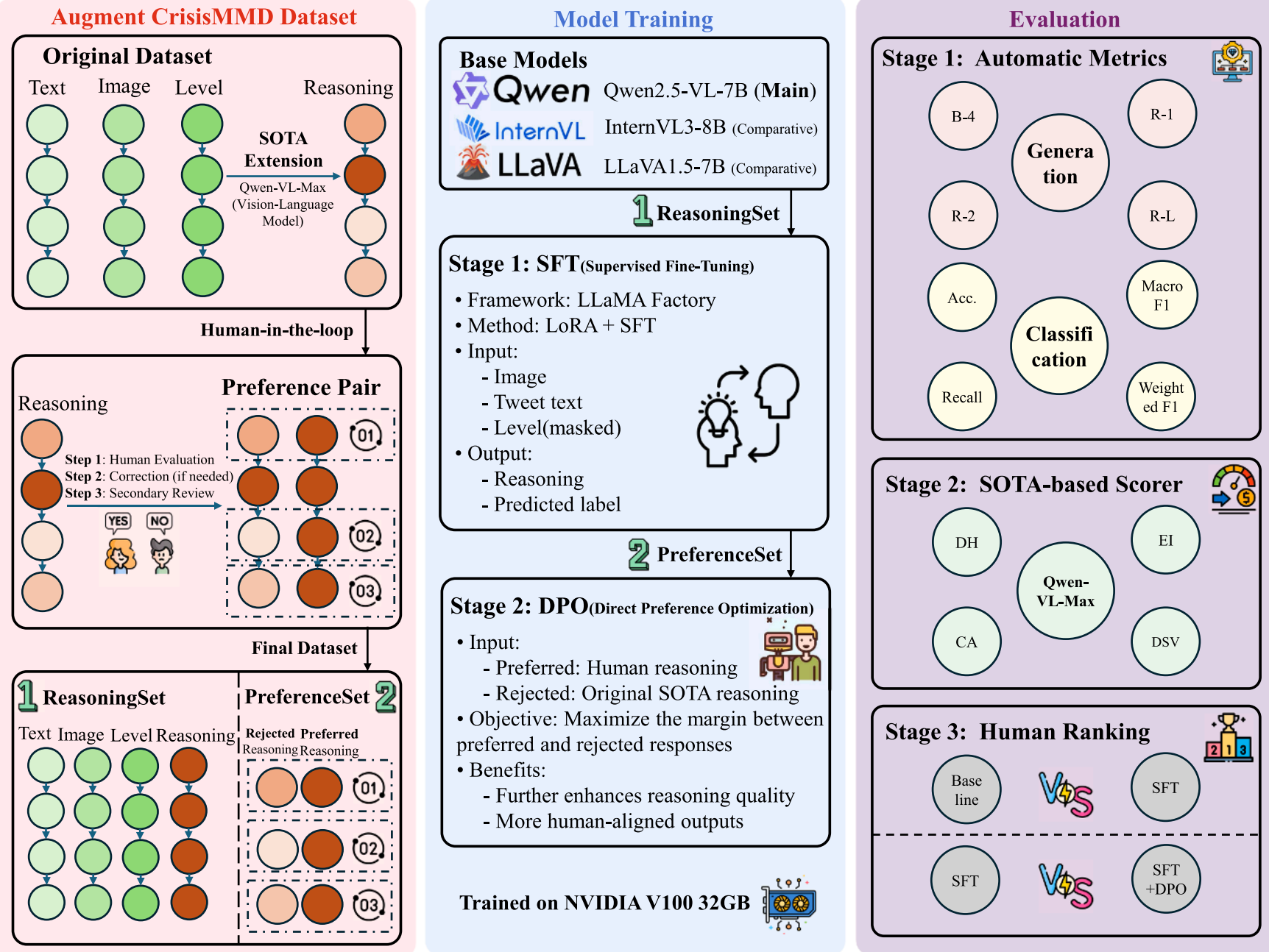

**Fig. 1.** Overview of the CrisisMMD-R framework, including (1) data construction and human-in-the-loop reasoning (left), (2) two-stage model training with Supervised Fine-Tuning and Direct Preference Optimization (middle), and (3) multi-level evaluation combining automatic metrics, model-based scoring, and human ranking (right). **Mottled/light-to-dark orange** indicates noisy model-generated reasoning drafts, while **solid orange** indicates final human-validated reasoning. Preference pairs are formed by matching each solid-orange (preferred) human-validated reasoning with the corresponding mottled/orange draft (rejected).

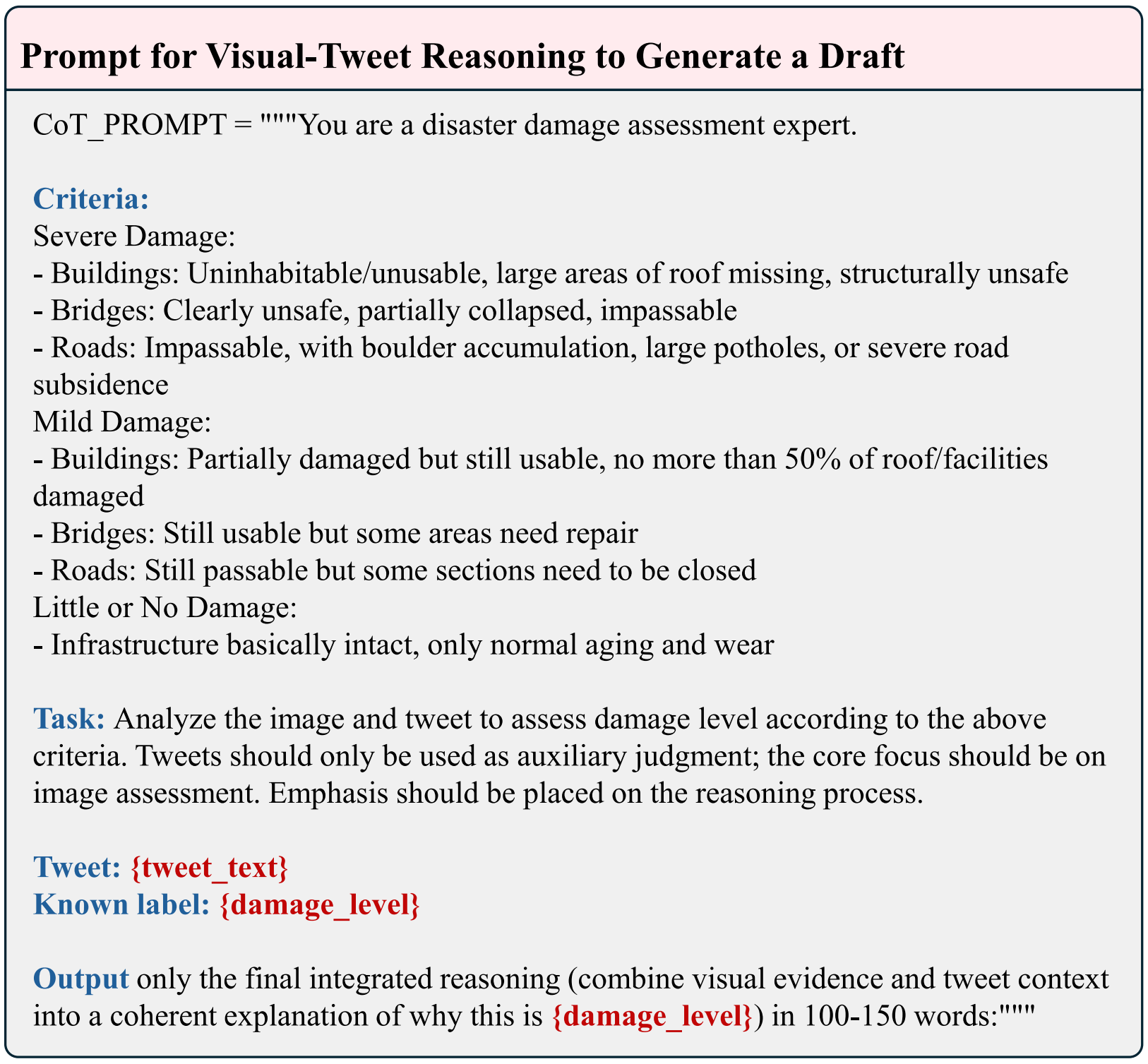

**Prompt for Visual-Tweet Reasoning to Generate a Draft**

CoT_PROMPT = """You are a disaster damage assessment expert.

**Criteria:**
Severe Damage:
- Buildings: Uninhabitable/unusable, large areas of roof missing, structurally unsafe
- Bridges: Clearly unsafe, partially collapsed, impassable
- Roads: Impassable, with boulder accumulation, large potholes, or severe road subsidence
Mild Damage:
- Buildings: Partially damaged but still usable, no more than 50% of roof/facilities damaged
- Bridges: Still usable but some areas need repair
- Roads: Still passable but some sections need to be closed
Little or No Damage:
- Infrastructure basically intact, only normal aging and wear

**Task:** Analyze the image and tweet to assess damage level according to the above criteria. Tweets should only be used as auxiliary judgment; the core focus should be on image assessment. Emphasis should be placed on the reasoning process.

**Tweet: {tweet_text}**
**Known label: {damage_level}**

**Output** only the final integrated reasoning (combine visual evidence and tweet context into a coherent explanation of why this is **{damage_level}**) in 100-150 words:"""

**Fig. 2.** Prompt for visual-tweet reasoning to generate a draft.

of explanations and conduct multiple prompt-engineering trials without incurring prohibitive expenses, ultimately yielding higher-quality prompts and data.

### 3.1.4. *Contributor's annotation phase*

Each draft is evaluated by a trained contributor who is given the image, the associated tweet, and the model's explanation. Contributors rate the draft along three dimensions; each dimension is scored from 1 to 5 according to the evaluation framework in Table 2. The final contributor decision is then passed to the professional review stage. If the *Total* score $\geq 12$ and no single dimension is $< 3$, the draft is "Approved". The evaluation framework, summarized in Table 2, is grounded in three key dimensions inspired by prior research [65–67] on multimodal reasoning and text evaluation. First, *Evidence Recognition* emphasizes that valid explanations must accurately identify and utilize the salient information present in the input, aligning with the principles outlined in the ONEEVAL benchmark (CAQA subset), which spans tasks “from basic evidence recognition to complex multi-hop reasoning” [65]. Second, *Reasoning Chain* assesses the logical consistency of the explanation, requiring that each step derive the conclusion through coherent and well-supported inference, as emphasized by Prasad et al. [66]. Third, *Text Naturalness* focuses on fluency, readability, and overall human-likeness of the generated explanation. Following contemporary text-generation evaluation standards [67], this dimension ensures that rationales are expressed clearly and naturally while explicitly indicating the assigned disaster damage level. Collectively these three dimensions provide a structured foundation for assessing both the factual grounding and linguistic quality of human-validated and model-generated explanations.

**Table 2**
Contributor evaluation framework: evaluation dimensions and corresponding score criteria (1–5).

| Dimension | Description | Criterion | Score |
|---|---|---|---|
| **Evidence Recognition** | Evaluates whether the model identifies salient visual and textual evidence relevant to disaster damage. | 1.1 Severely misses key evidence or misinterprets image/tweet content. | 1 |
| | | 1.2 Identifies only secondary evidence; multiple important details are missing. | 2 |
| | | 1.3 Captures main evidence but overlooks some important information. | 3 |
| | | 1.4 Finds most important evidence; only minor omissions that do not affect reasoning. | 4 |
| | | 1.5 Accurately identifies all key visual elements and crucial tweet information. | 5 |
| **Reasoning Chain** | Assesses logical soundness and evidential support from evidence to conclusion. | 2.1 No coherent logic; conclusion disconnected from evidence. | 1 |
| | | 2.2 Broken chain of reasoning with large unsupported jumps. | 2 |
| | | 2.3 Direction is visible, but several links are weak or under-supported. | 3 |
| | | 2.4 Mostly sound logic with occasional small gaps; still comprehensible. | 4 |
| | | 2.5 Clear causal chain with all steps well supported by evidence. | 5 |
| **Text Naturalness** | Measures clarity, fluency, and human-likeness, including explicit statement of the final damage level. | 3.1 Fragmented or repetitive language that is difficult to read. | 1 |
| | | 3.2 Strongly “AI-like” tone; lacks natural transitions; includes unexplained markers/symbols (tweet quotes excepted). | 2 |
| | | 3.3 Clearly AI-generated but of acceptable readability. | 3 |
| | | 3.4 Generally fluent and clear with only occasional AI traces. | 4 |
| | | 3.5 Reads like a human expert analysis **and** explicitly states one of *Little or no damage / Mild damage / Severe damage*. | 5 |

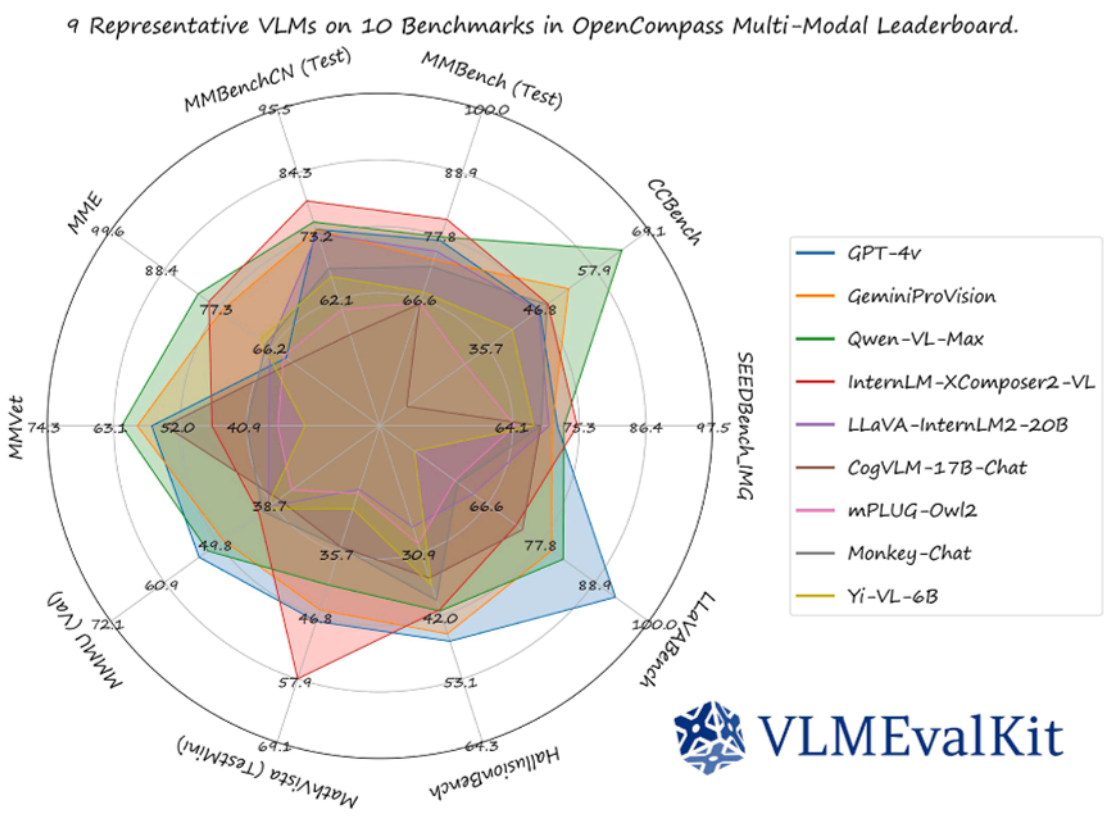

**Fig. 3.** OpenCompass multimodal leaderboard [64].

The quantitative thresholds defined in Table 2 directly determine the branching of each draft explanation into either the ReasoningSet or the PreferenceSet. Specifically, drafts with a total score $\geq 12$ and no individual dimension score below 3 are provisionally accepted by contributors and forwarded for rapid reviewer verification. Drafts failing to meet this threshold enter the correction workflow and are later paired with their revised versions to form preference pairs. Out of the 3526 initial machine-generated drafts 1571 has been filtered by contributors, 1284 explanations satisfied both contributor and reviewer criteria and were consolidated into the final ReasoningSet, yielding an overall acceptance rate of 81.7%. The remaining 287 drafts requiring correction were selected to construct high-quality rejected-preferred pairs for DPO, forming the PreferenceSet. The remaining corrected drafts were excluded from alignment training to maintain strict separation between supervised and preference-based supervision signals. Contributors in the HITL pipeline were assigned disjoint subsets of items, and therefore, no instance received multiple independent contributor ratings. Consequently, classical inter-annotator agreement measures such as Krippendorff’s $\alpha$ or Fleiss’ $\kappa$ cannot be directly computed, as they require overlapping annotations on identical samples. This annotation strategy was intentionally adopted to scale annotation throughput while reserving expert effort for validation and correction stages. To ensure reliability under single-annotation conditions, we adopted a hierarchical validation protocol in which professional reviewers verified high-confidence items and revised lower-confidence explanations. Such adjudication-based designs are commonly used in dataset construction when expert validation is considered a stronger reliability signal than redundant annotation [68–70]. In these settings, reliability is established through a calibrated review stage rather than contributor agreement alone.

Specifically, contributor scores exceeding the acceptance threshold were verified, while uncertain cases were rewritten to form preference pairs. This transforms potential contributor variability into supervised correction signals rather than propagating noise into the final dataset. The reviewer stage, therefore, acts as a quality filter that converts single-pass annotations into validated supervision, improving label consistency without requiring redundant labeling. Such two-stage annotation pipelines have been shown to produce higher quality corpora compared to majority voting among non-expert annotators, particularly for complex semantic tasks where expert adjudication better approximates ground truth [69,70]. Accordingly, reliability in our dataset derives from calibrated contributor training and systematic expert correction rather than pairwise agreement among contributors.

#### *3.1.5. Professional review phase*

Reviewers examine every entry and follow two distinct paths depending on the contributor’s decision. For contributor-approved items,

**Table 3**
Verification decisions by reviewer.

| | Accept | Revise |
|---|---|---|
| Reviewer 1 | 642 | 144 |
| Reviewer 2 | 642 | 143 |

they conduct a quick yet comprehensive verification across the three target dimensions shown in Table 2. Entries that pass both the contributor and reviewer checks are consolidated into the SFT dataset. For contributor-rejected items, reviewers revise the content guided by the contributor's comments to obtain a corrected version. This human editing process inherently encodes human preferences; by pairing the pre- and post-revision versions, we obtain preference pairs that constitute our DPO dataset.

As mentioned in Section 3.1.4, traditional inter-annotator agreement metrics such as Fleiss' $\kappa$ or Krippendorff's $\alpha$ cannot be computed, which requires multiple independent judgments on identical instances to estimate chance-corrected agreements [71,72]. In adjudication-based annotation workflows commonly used in safety-critical and clinical AI datasets, reliability is instead assessed at the process level rather than the item level. The key question becomes whether outcomes depend on the evaluator or on the evaluation protocol. Therefore, we model reliability as a statistical independence problem between reviewer identity and decision outcome (Accept / Revise). Here, reviewer identity refers only to which assigned expert performed the verification step and not to personal attributes. If decisions are independent of reviewer identity, then the review protocol is stable and reproducible. If dependent, systematic reviewer bias exists. This approach follows reliability evaluation practices in HITL validation pipelines, where single-adjudicator review replaces multi-coder agreement, and consistency is verified through bias-detection statistics rather than agreement coefficients [73]. Accordingly, we evaluate process reliability using a chi-square test of independence ($\chi^2$):

We construct a $2 \times 2$ Contingency Table 3. It uses reviewer identity (Reviewer 1, Reviewer 2) and verification outcome (Accept, Revise) as categorical variables.

The null hypothesis is:

$H_0$ : Reviewer identity and decision outcome are independent

The alternative hypothesis is:

$H_a$ : Reviewer identity influences decision outcome

We apply Pearson's Chi-square test of independence ($\chi^2$). For a contingency table with $r$ rows and $c$ columns, the expected frequency in each cell is:

$$E_{ij} = \frac{R_i C_j}{N}$$

where $R_i$ is the row total, $C_j$ is the column total, and $N$ is the total number of observations.

The test statistic is computed as:

$$\chi^2 = \sum_{i=1}^{r} \sum_{j=1}^{c} \frac{(O_{ij} - E_{ij})^2}{E_{ij}}$$

with degrees of freedom:

$$df = (r-1)(c-1)$$

For the $2 \times 2$ table, $df = 1$.

Using the observed counts, the test produced:

$$\chi^2(1) = 0.0014, \quad p = 0.97$$

Since $p \gg 0.05$, we fail to reject the null hypothesis. Therefore, verification outcomes are statistically independent of reviewer identity. This indicates that both reviewers applied the evaluation criteria consistently and that acceptance or revision decisions are not attributable to individual evaluator bias. Consequently, the professional review stage demonstrates procedural reliability even in the absence of overlapping annotations.

**Table 4**
SFT-dataset split and damage-level distribution.

| | Train | Dev | Test | Total |
|---|---|---|---|---|
| Little or No Damage | 91 | 12 | 11 | 114 |
| Mild Damage | 170 | 21 | 22 | 213 |
| Severe Damage | 766 | 95 | 96 | 957 |
| Total | 1027 | 128 | 129 | 1284 |

**Table 5**
DPO-dataset split and damage-level distribution.

| | Train | Dev | Test | Total |
|---|---|---|---|---|
| Little or No Damage | 34 | 4 | 5 | 43 |
| Mild Damage | 74 | 10 | 9 | 93 |
| Severe Damage | 121 | 15 | 15 | 151 |
| Total | 229 | 29 | 29 | 287 |

**Table 6**
Profiles of contributors and professional reviewers involved in dataset curation.

| Annotator ID | Role | Education | Crisis-domain publications |
|---|---|---|---|
| Contributor 1 | Contributor | Master | No |
| Contributor 2 | Contributor | Master | Yes |
| Contributor 3 | Contributor | Master | Yes |
| Reviewer 1 | Professional reviewer | PhD | Yes |
| Reviewer 2 | Professional reviewer | PhD | Yes |

### 3.1.6. *Final datasets*

The final dataset composition is a direct outcome of the scoring dimensions and decision thresholds defined in Table 2. Explanations that met the contributor threshold (Total $\geq 12$ with no dimension $< 3$) and passed reviewer verification were incorporated into the ReasoningSet. This process resulted in 1284 validated rationales, which serve as the supervision source for LoRA-based supervised fine-tuning. Conversely, drafts that failed the contributor threshold entered the correction workflow. For a subset of 287 such cases, both the original rejected explanation and its human-corrected revision were retained to form explicit rejected-preferred pairs. These constitute the PreferenceSet used for DPO. The remaining corrected entries were excluded from DPO construction to prevent overlap between supervised and preference-aligned data.

Table 6 summarizes the contributor and professional reviewer profiles involved in this curation process. Contributors were selected based on familiarity with disaster damage assessment and proficiency in English social-media discourse, while professional reviewers were selected for domain expertise and experience in multimodal data annotation and quality control. All participants completed task-specific training covering the CrisisMMD damage-severity criteria, the three-dimensional evaluation rubric, and calibration on representative examples prior to formal annotation and review.

The final dataset consists of two complementary components: the ReasoningSet and the PreferenceSet. The data are divided into training, validation, and test splits following an 80, 10, and 10 percent ratio, with detailed class distribution provided in Table 4. To mitigate potential sampling bias under the highly skewed label distribution, we verified that the split preserves the class proportions across subsets (Table 4). We do not apply any rebalancing strategies and instead train on the natural distribution. Given the dominance of Severe damage, we later report imbalance- and risk-aware metrics beyond accuracy (see Section 4.1). The PreferenceSet, on the other hand, comprises 287 paired entries derived from the same set of items, where each pair includes a rejected rationale generated by Qwen-VL-Max that did not pass human review and a corresponding preferred rationale refined by human annotators. We follow the same 80/10/10 splitting protocol for the PreferenceSet and similarly confirm that its class distribution is preserved across train/dev/test (Table 5). A detailed breakdown of the PreferenceSet composition and damage-level distribution is provided in Table 5.

### 3.2. Problem definition

Given an image $I$ and its accompanying tweet $T$, the model must generate

$$< \hat{Y},\ \hat{R} >= f_{\theta}(I, T) \tag{1}$$

where $\hat{Y} \in \mathcal{Y} = \{\text{LITTLE OR NO}, \text{MILD}, \text{SEVERE}\}$ the predicted damage label and $\hat{R}$ is a free-form textual rationale usually written in English. The two outputs are produced in a single autoregressive decoding pass to avoid any post-hoc explanation, i.e., generating the label first and subsequently fabricating the reasoning. Instead, binding prediction and explanation in the same decoding process allows the model to jointly optimize for correctness and interpretability.

### 3.3. Base architecture

The proposed framework is primarily implemented using Qwen2.5-VL-7B [23], an open-weight vision-language model that adopts a ViT encoder, MLP merger, and LLM decoder architecture. To ensure generalizability of the dataset and training pipeline, the same two-stage fine-tuning and alignment procedures are also applied to two representative open-source models, InternVL3-8B [24], and LLaVA1.5-7B [25]. Although these models differ in backbone design and alignment mechanisms, they share the encoder-decoder paradigm, making them suitable for cross-validation and comparative analysis. Unless otherwise stated, the architectural description focuses on Qwen2.5-VL-7B, which serves as the primary model in this study.

The vision encoder consists of a Vision Transformer (ViT) with 32 transformer blocks employing two-dimensional rotary positional embedding (2D-RoPE) and window attention. Input images are resized so that their height and width are multiples of 28, then divided into $14{\times}14$ patches with a stride of 14. Window attention with a window size of 112 is applied in most layers, while full attention is activated at layers $7, 15, 23, 31$ to ensure global feature interaction. Each image patch is embedded into a 1280-dimensional representation. To maintain computational efficiency in the language decoder, four spatially adjacent patch embeddings are concatenated and passed through a two-layer multilayer perceptron (MLP) merger. This operation reduces the visual sequence length by a factor of four while aligning the visual representation with the dimensionality of the language model's token embeddings.

The language decoder builds upon the pretrained Qwen2.5 7B checkpoint but replaces the original one-dimensional rotary position embedding with Multimodal RoPE (M-RoPE), which aligns both textual and visual tokens within a coherent positional framework. The decoder comprises 28 transformer layers, each with 28 attention heads of dimension 128, and employs key/value sharing across four KV heads to improve memory efficiency. The model has been trained on 4.1 trillion mixed-modality tokens, providing extensive grounding for multimodal reasoning tasks. The detailed configuration of Qwen2.5-VL-7B, including key hyperparameters for its vision, merger, and language components, is summarized in Table 7.

**Table 7**
Configuration of Qwen2.5-VL-7B.

| Module | Configuration | Value |
|---|---|---|
| **Vision Transformer (ViT)** | Hidden Size | 1280 |
| | # Layers | 32 |
| | # Num Heads | 16 |
| | Intermediate Size | 3456 |
| | Patch Size | 14 |
| | Window Size | 112 |
| | Full Attention Block Indexes | {7,15,23,31} |
| **Vision-Language Merger** | In Channel | 1280 |
| | Out Channel | 3584 |
| **Large Language Model (LLM)** | Hidden Size | 3584 |
| | # Layers | 28 |
| | # KV Heads | 4 |
| | Head Size | 128 |
| | Intermediate Size | 18,944 |
| | Embedding Tying | × |
| | Vocabulary Size | 151,646 |
| | # Trained Tokens | 4.1T |

### 3.4. Prompt design and data serialization

For both stages of model adaptation, Supervised Fine-Tuning (SFT) and Direct Preference Optimization (DPO), the raw image-tweet pairs are serialized into a structured JSON corpus compatible with the `LLaMA-Factory` toolkit. The framework supports two native schemas, ALPACA and SHAREGPT; however, for multimodal dialogue applications, the latter is explicitly recommended by the authors, and thus the SHAREGPT layout is consistently adopted throughout this study [74]. The process begins with a fixed system prompt that defines the instruction template for all experiments. As illustrated in Fig. 4, this expert-level instruction outlines the official damage assessment criteria, emphasizes that the image serves as the primary evidence source, and constrains the generated rationale to 100–150 words. The system prompt is conceptually aligned with the design presented in Section 3.1.3, following the same principles of evidence-based assessment and multimodal reasoning. This consistency ensures that both automated and human-curated rationales adhere to a uniform interpretive framework during dataset construction and fine-tuning.

During the SFT stage, each training instance, represented as $\langle \text{image}, \text{tweet}, \text{system_prompt}, \text{gold rationale} \rangle$, is serialized into a SHAREGPT record in which the `<image>` token instructs the `LLaMA-Factory` to attach visual embeddings. The human-refined explanation is stored in the `final_response` field, and the `system_prompt` corresponds to the instruction template described above. An example of this format is provided in Listing 1.

In the DPO stage, serialization follows a similar structure but incorporates an additional preference dimension. Each training instance, denoted as $\langle \text{image}, \text{tweet}, \text{system_prompt}, \text{chosen}, \text{rejected} \rangle$, contains two possible rationales: a human-approved `chosen` rationale and a machine-generated `rejected` rationale. Both are stored under separate fields while maintaining the same `<image>` token and `system_prompt` schema. This design enables the model to explicitly learn preference relationships, aligning its reasoning with human judgment. The corresponding implementation is shown in Listing 2.

To prevent label leakage, the ground-truth damage label is never included in the SHAREGPT input during SFT/DPO training or evaluation.

### 3.5. Stage I: Supervised fine-tuning (SFT)

The supervised fine-tuning (SFT) stage is primarily applied to Qwen2.5-VL-7B [23], with comparable experiments conducted on InternVL3-8B [24] and LLaVA1.5-7B [25] to validate the generality of the proposed framework. The training utilizes the REASONINGSET, comprising 1027 instances (80% of the dataset) for training and 128 instances (10%) for validation, while dataset composition and class distribution are detailed in Section 3.1.6 and Table 4. To enable parameter-efficient adaptation, the fine-tuning process employs the Low-Rank Adaptation (LoRA) technique [62], where all linear layers are targeted with a rank of $r = 16$ and a scaling factor $\alpha = 32$, while the vision encoder and multimodal projector remain frozen.

**You are a disaster damage assessment expert.**

*Criteria*

**Severe Damage** - Buildings: Uninhabitable/unusable, large areas of roof missing, structurally unsafe - Bridges: Clearly unsafe, partially collapsed, impassable - Roads: Impassable, with boulder accumulation, large potholes, or severe road subsidence

**Mild Damage** - Buildings: Partially damaged but still usable, no more than 50% of roof/facilities damaged - Bridges: Still usable but some areas need repair - Roads: Still passable but some sections need to be closed

**Little or No Damage** - Infrastructure basically intact, only normal aging and wear

*Task*

Analyze the image and tweet to assess damage level according to the above criteria. Tweets should only be used as auxiliary judgment; the core focus should be on image assessment. Emphasis should be placed on the reasoning process. Provide integrated reasoning that combines visual evidence and tweet context into a coherent explanation in 100-150 words.

**Fig. 4.** Fixed system prompt used in all experiments.

```python
# Construct ShareGPT format in SFT stage
sample = {
    "conversations": [
        {
            "from": "human",
            "value": f"<image>{system_prompt}
                Tweet: '{row['tweet_text']}'"
        },
        {
            "from": "gpt",
            "value": row['final_response']
        }
    ],
    "images": [image_path]
}
```

**Listing 1.** SFT ShareGPT

The training objective follows the problem formulation in Section 3.2, in which the model jointly learns to generate both the damage label $\hat{Y}$ and the rationale $\hat{R}$ within a single autoregressive decoding pass. Formally, the corresponding loss function is expressed as

$$\mathcal{L}_{\text{SFT}} = -\sum_{t=1}^{|Y|+|R|} \log p_\theta(s_t \mid s_{<t}, I, T) \tag{2}$$

where $s = [Y, R]$ denotes the concatenated output sequence. This joint formulation ensures that prediction and explanation are learned concurrently, preventing post-hoc rationalization and promoting interpretability. The model is trained for three epochs using a batch size of eight, the AdamW optimizer, and a learning rate of $7 \times 10^{-5}$ under a cosine decay schedule. Model checkpoints are evaluated periodically, and the one achieving the lowest validation loss is selected for subsequent alignment.

### 3.6. Stage II: Direct preference optimization (DPO)

The second stage of training employs Direct Preference Optimization (DPO) to further align model outputs with human-preferred reasoning. As in the SFT stage, DPO is primarily conducted on Qwen2.5-VL-7B [23], while additional experiments on InternVL3-8B [24] and LLaVA1.5-7B [25] serve to validate the generality and robustness of the proposed pipeline. The training utilizes the PreferenceSet, which contains 229 instances (80% for training and 10% for validation). Each sample consists of a pair of structured responses $\hat{s} = [\hat{Y}, \hat{R}]$ (i.e., the joint output containing both the predicted damage label and the corresponding rationale), one chosen and one rejected, derived from the HITL annotation process described in Section 3.1.6 and summarized in Table 5. The damage label in each structured response is kept consistent with the original CrisisMMD ground-truth annotation.

```python
# Construct ShareGPT format in DPO stage
sample = {
    "conversations": [
        {
            "from": "human",
            "value": f"<image>{system_prompt}
                Tweet: '{row['tweet_text']}'"
        },
    ],
    "chosen": {
        "from": "gpt",
        "value": row['modified_response']
    },
    "rejected": {
        "from": "gpt",
        "value": row['final_response']
    },
    "images": [image_path]
}
```

**Listing 2.** DPO ShareGPT

Parameter-efficient adaptation is achieved using the Low-Rank Adaptation (LoRA) technique [62], adopting the same configuration as in the SFT stage, where all linear layers are targeted with a rank of $r = 16$ and a scaling factor $\alpha = 32$, while the vision encoder and multimodal projector remain frozen. The model is optimized to assign a higher likelihood to preferred responses over rejected ones for a given image $I$ and tweet $T$, according to the objective

$$\mathcal{L}_{\text{DPO}} = -\log \sigma\left(\frac{1}{\tau}\left(\log p_\theta(\hat{s}_c \mid I, T) - \log p_\theta(\hat{s}_r \mid I, T)\right)\right) \tag{3}$$

where $\tau$ denotes the temperature parameter, controlling the sharpness of preference weighting. $\hat{s}_c = [\hat{Y}_c, \hat{R}_c]$ and $\hat{s}_r = [\hat{Y}_r, \hat{R}_r]$ denote the chosen and rejected structured responses, respectively.

The training runs for four epochs with a batch size of eight, using the AdamW optimizer with a learning rate of $5 \times 10^{-6}$ under a cosine decay schedule. Model checkpoints are evaluated periodically, and the one achieving the lowest validation loss is retained for deployment. The resulting DPO-aligned models produce responses that better reflect human judgment, enhancing interpretability, trustworthiness, and practical utility in reliability-oriented decision support systems.

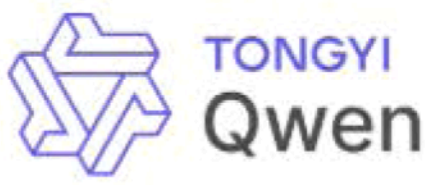

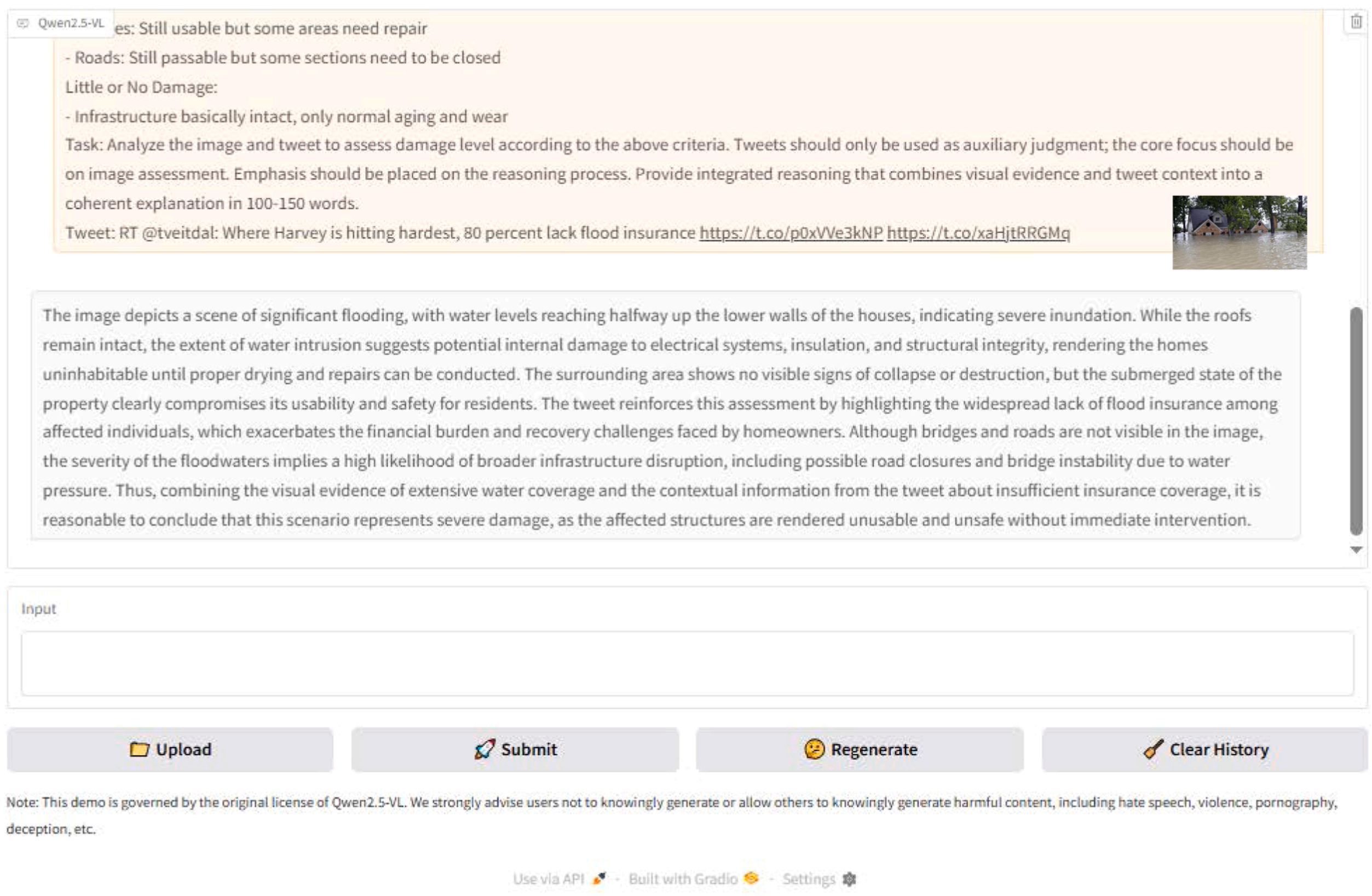

**Fig. 5.** Gradio-based prototype of the system demonstrating real-time explainable disaster damage assessment. The interface accepts tweet-image pairs as input, outputs both the predicted damage level and the corresponding reasoning for reliability-oriented decision support. An API access endpoint is also provided at the bottom of the interface for system integration.

### *3.7. Prototype deployment for reliability-oriented decision support*

To demonstrate the practical usability of the trained explainable VLMs, we implement a lightweight prototype interface using the Gradio framework. As shown in Fig. 5, the interface accepts a tweet-image pair and renders both the predicted damage level and the model's natural-language rationale under the assessment criteria. The prototype enables real-time interaction and transparent inspection of evidence, supporting human-in-the-loop verification under time pressure. Deployment is flexible, and the service can run on a private, access-controlled server for sensitive data, or be exposed via a public endpoint when appropriate; programmatic access is provided via the gradio_client library, which issues HTTPS requests to named app endpoints for integration with crisis dashboards and command-and-control systems. From a reliability perspective, the interface operationalizes our pipeline by turning model outputs into auditable, reviewable artifacts that can be embedded into command-and-control workflows and crisis dashboards. This step bridges the methodological pipeline in Fig. 1 with operational use, paving a path toward continuous, risk-informed decision support.

## 4. Results and discussions

This section presents the evaluation results of our two-stage training approach. We first describe our comprehensive evaluation framework that combines automatic metrics, SOTA-based assessment, and human ranking, as illustrated in the Evaluation part of Fig. 1. Table 8 summarizes the overall evaluation results across all stages and metrics, with detailed analysis and discussion provided in the following sub-sections.

### *4.1. Evaluation framework*

To comprehensively assess model performance, a three-tiered evaluation framework, that integrates automatic metrics, state-of-the-art (SOTA) model-based assessment, and human ranking, is employed. The first tier focuses on automatic metrics encompassing both classification and generation measures, as summarized in Table 8. Classification performance is evaluated using Accuracy, which represents the overall correctness rate; Macro F1, the average of per-class F1 scores that assigns equal weight to all categories and is particularly valuable for imbalanced datasets where severe damage cases dominate; and Per-class F1, which measures the model's performance on individual classes such as Little/No, Mild, and Severe damage, highlighting improvements in minority categories. Beyond accuracy-style metrics, we further report a cost-weighted, risk-aware error measure to reflect the asymmetric operational risk of underestimating damage (e.g., Severe→Little/No is more harmful than Mild↔Little/No). Specifically, we map the ordered damage levels to $\{0, 1, 2\}$ corresponding to {Little/No, Mild, Severe}, and define an asymmetric cost function for each prediction:

$$c(y, \hat{y}) = \begin{cases} 0, & y = \hat{y}, \\ w_{\text{under}} \cdot (y - \hat{y}), & \hat{y} < y, \\ w_{\text{over}} \cdot (\hat{y} - y), & \hat{y} > y, \end{cases} \tag{4}$$

where $w_{\text{under}}$ penalizes underestimation (riskier) and $w_{\text{over}}$ penalizes overestimation. The overall risk is then computed as the expected cost over the test set:

$$\text{Risk} = \frac{1}{N} \sum_{i=1}^{N} c(y_i, \hat{y}_i), \tag{5}$$

which can be equivalently derived from confusion matrices. In our main setting we fix $w_{\text{under}} = 5$ and conduct sensitivity analysis with $w_{\text{over}} \in \{2, 3, 4\}$ to verify that the conclusions are stable under different relative penalties for overestimation. To complement these, generation metrics including BLEU-4 and ROUGE (ROUGE-1, ROUGE-2, ROUGE-L) are used to assess the n-gram overlap between generated and reference explanations, thereby capturing different aspects of rationale fluency and coherence. This dual evaluation enables the simultaneous measurement of classification accuracy and explanation quality.

**Table 8**
Comprehensive evaluation results. Within each backbone, the best SFT scores are in **bold**, while the best DPO scores are underlined.

| Backbone | Method | Dataset | Overall↑ | | Per-class F1↑ | | | Generation↑ | | | | SOTA-based Scorer↑ | | | |
|---|---|---|---|---|---|---|---|---|---|---|---|---|---|---|---|
| | | | Acc. | M-F1 | Little/No | Mild | Severe | BLEU-4 | R-1 | R-2 | R-L | DH | EI | CA | DSV |
| Qwen2.5-VL 7B-Instruct | Baseline | SFT-Testset | 73.64 | 44.46 | 32.00 | 15.38 | 85.99 | 56.29 | 42.86 | 15.82 | 27.69 | 3.91 | 3.01 | 3.89 | 3.75 |
| | SFT | SFT-Testset | **78.29** | **57.42** | **33.33** | **48.78** | **90.16** | **68.47** | **54.03** | **29.44** | **39.42** | **4.72** | **4.20** | **4.68** | **4.60** |
| | SFT | DPO-Testset | 58.62 | 52.67 | 50.00 | 35.29 | 72.73 | 48.60 | 40.70 | 15.92 | 28.59 | 4.34 | 4.31 | 4.38 | 4.28 |
| | SFT + DPO | DPO-Testset | 55.17 | 52.21 | 50.00 | 42.11 | 64.52 | 54.34 | 39.31 | 13.11 | 24.50 | 4.59 | 4.52 | 4.38 | 4.41 |
| InternVL 3-8B | Baseline | SFT-Testset | 72.87 | 48.52 | **30.00** | 29.27 | 86.29 | 58.56 | 41.98 | 14.48 | 25.94 | 4.10 | 3.88 | 4.26 | 4.00 |
| | SFT | SFT-Testset | **76.74** | **55.64** | 28.57 | **48.98** | **89.36** | **68.51** | **54.27** | **28.92** | **39.57** | **4.64** | **4.48** | **4.60** | **4.57** |
| | SFT | DPO-Testset | 55.17 | 46.48 | 28.57 | 42.11 | 68.75 | 48.63 | 40.75 | 16.32 | 29.10 | 4.38 | 4.34 | 4.21 | 3.97 |
| | SFT + DPO | DPO-Testset | 55.17 | 52.36 | 44.44 | 52.63 | 60.00 | 56.10 | 39.65 | 12.77 | 24.50 | 4.48 | 4.45 | 4.24 | 4.00 |
| LLaVA 1.5-7B | Baseline | SFT-Testset | **75.19** | **52.17** | **43.48** | 27.59 | 85.44 | 46.57 | 34.89 | 10.36 | 24.24 | 3.78 | 3.48 | 3.53 | 3.54 |
| | SFT | SFT-Testset | 74.42 | 50.61 | 27.27 | **35.56** | **89.01** | **67.59** | **52.91** | **27.79** | **38.15** | **4.72** | **4.47** | **4.46** | **4.52** |
| | SFT | DPO-Testset | 75.86 | 71.90 | 66.67 | 66.67 | 82.35 | 49.26 | 41.52 | 15.04 | 28.71 | 4.45 | 4.41 | 4.14 | 4.14 |
| | SFT + DPO | DPO-Testset | 55.17 | 52.21 | 50.00 | 42.11 | 64.52 | 56.68 | 38.48 | 12.96 | 23.71 | 4.48 | 4.34 | 4.17 | 4.17 |

The second tier employs a SOTA-based scorer using Qwen-VL-Max to evaluate the practical value of generated explanations across four crisis-specific dimensions: Diagnostic Helpfulness, which assesses how well explanations justify the assigned damage level and connect visual evidence to the decision; Evidence Integration, which evaluates the synthesis of textual and visual cues to form a coherent interpretation; Contextual Awareness, which examines the model's understanding of disaster-type-specific characteristics such as floods, fires, and hurricanes; and Decision Support Value, which measures the actionable utility of the explanation for crisis management and resource planning. Each dimension is scored on a five-point scale, with higher scores indicating stronger performance. The scoring process follows the standardized evaluation prompt illustrated in Fig. 6. This SOTA-based assessment complements conventional metrics by capturing operational relevance and decision-support potential, aspects that purely linguistic metrics cannot fully represent. The Qwen-VL-Max judge is provided with the tweet text, the input image, and the model outputs (predicted label + rationale), and is blind to CrisisMMD ground-truth annotations (ground-truth labels/rationales are not shown). To verify that our conclusions are not an artifact of a specific judge model, we repeat the same scoring protocol with an alternative judge (GPT-4o-mini [75]) and observe consistent relative rankings across backbones and test sets. For readability, we report the full results in Appendix A (Table A.14).

The third tier incorporates human ranking to address the limitations of automatic and model-based evaluations. Recognizing that metric-based evaluations often fail to capture nuanced differences between model output, human judgments are collected through pairwise comparison methods that provide more reliable and interpretable results. Prior studies have shown that pairwise and ranking-based methods, such as Best-Worst Scaling [76] and ranking-based magnitude estimation [77], achieve higher inter-annotator agreement than absolute scoring. These approaches are particularly suitable for evaluating fine-grained quality differences arising from the two-stage training process, where subtle variations in explanation clarity, coherence, or reasoning depth can significantly impact perceived quality. Together, these three complementary tiers, automatic metrics, SOTA-based evaluation, and human ranking, form a comprehensive, multi-perspective framework for evaluating both the predictive accuracy and interpretability of the proposed multimodal disaster assessment models.

### 4.2. Supervised fine-tuning (SFT) results

The SFT stage demonstrates substantial improvements across all evaluation metrics, as shown in Table 8. We analyze these results from three perspectives: classification performance, generation quality, and practical implications.

#### 4.2.1. Classification performance

The supervised fine-tuning (SFT) stage demonstrates consistent and substantial improvements across all evaluation dimensions, as presented in Table 8. From the perspective of classification performance, SFT markedly enhances both accuracy and Macro F1 across all three backbones, with the most significant improvements observed in the under-represented Mild damage category. For Qwen2.5-VL-7B, overall accuracy increases from 73.64% to 78.29%, while Macro F1 rises from 44.46% to 57.42%, representing a 29.1% relative improvement. Per-class F1 shows modest gains for Little/No damage (32.00% to 33.33%) and solid improvement for Severe (85.99% to 90.16%), but a dramatic jump for Mild damage from 15.38% to 48.78%, corresponding to a 217% relative gain. These results confirm that SFT effectively mitigates the model's tendency to confuse moderate with severe damage, a key challenge in multimodal disaster assessment. InternVL3-8B follows a similar trend, with accuracy increasing from 72.87% to 76.74% and Macro F1 from 48.52% to 55.64% (a 14.7% relative gain), again driven largely by improved recognition of Mild damage cases (29.27% to 48.98%). Although LLaVA1.5-7B shows more modest changes, with accuracy slightly decreasing from 75.19% to 74.42%, it still benefits from targeted improvements in the Mild (27.59% to 35.56%) and Severe (85.44% to 89.01%) categories.

**You are a disaster damage assessment evaluator.**

*Task*
Please evaluate the following disaster damage assessment based on the tweet, image, and model output.

*Scoring Dimensions (1-5 scale, where 5 is best)*
**1. Diagnostic_helpfulness:** How well does the explanation help understand why this specific damage level was assigned? Does it clearly connect visual evidence to the classification?
**2. Evidence_integration:** How effectively does the output integrate and synthesize information from both the image and tweet text? Does it demonstrate understanding of multiple evidence sources?
**3. Contextual_awareness:** Does the output show understanding of the specific disaster type (flood/fire/hurricane) and its characteristic damage patterns? Is domain knowledge appropriately applied?
**4. Decision_support_value:** How useful would this assessment be for crisis response teams? Does it provide actionable insights for resource allocation and response planning?

*Output Format*
**IMPORTANT:** Return ONLY the JSON object below without any additional text or explanation:

```
{
  "scores": {
    "diagnostic_helpfulness": <score>,
    "evidence_integration": <score>,
    "contextual_awareness": <score>,
    "decision_support_value": <score>
  }
}
```

**Fig. 6.** Fixed scoring prompt used for evaluation experiments.

**Table 9**
Confusion matrices for baseline and SFT models across three backbones.

(a) Qwen2.5-VL-7B

| Baseline | | Predicted | | | SFT | Predicted | | |
|---|---|---|---|---|---|---|---|---|
| | | L/N | Mild | Severe | | L/N | Mild | Severe |
| Actual | L/N | 4 | 2 | 5 | L/N | 4 | 6 | 1 |
| | Mild | 3 | 2 | 17 | Mild | 3 | 10 | 9 |
| | Severe | 7 | 0 | 89 | Severe | 6 | 3 | 87 |

(b) InternVL3-8B

| Baseline | | Predicted | | | SFT | Predicted | | |
|---|---|---|---|---|---|---|---|---|
| | | L/N | Mild | Severe | | L/N | Mild | Severe |
| Actual | L/N | 3 | 6 | 2 | L/N | 3 | 8 | 0 |
| | Mild | 2 | 6 | 14 | Mild | 2 | 12 | 8 |
| | Severe | 4 | 7 | 85 | Severe | 5 | 7 | 84 |

(l) LLaVA1.5-7B

| Baseline | | Predicted | | | SFT | Predicted | | |
|---|---|---|---|---|---|---|---|---|
| | | L/N | Mild | Severe | | L/N | Mild | Severe |
| Actual | L/N | 5 | 1 | 5 | L/N | 3 | 7 | 1 |
| | Mild | 1 | 4 | 17 | Mild | 5 | 8 | 9 |
| | Severe | 6 | 2 | 88 | Severe | 3 | 8 | 85 |

The confusion matrices for all three backbones (Table 9) further illustrate this effect. Taking Qwen2.5-VL-7B as an example, the baseline severely misclassifies Mild damage, labeling 77.3% of these cases as Severe. After SFT, the correct classification of Mild cases increases from 2 to 10 (45.5%), while misclassifications as Severe drop from 17 to 9 (40.9%). This demonstrates that SFT improves the model's calibration, particularly in recognizing subtle category boundaries and preventing overestimation. In operational contexts, such calibration is essential, as baseline models often overpredict severity, potentially leading to unnecessary alarms or misallocation of resources. The SFT-adjusted model corrects this bias, improving reliability and decision alignment, and Table 10 further quantifies this reliability improvement using an asymmetric, cost-weighted risk metric that penalizes underestimation more heavily ($w_{\text{under}} = 5$) while sweeping $w_{\text{over}} \in \{2, 3, 4\}$ for sensitivity analysis. Under this risk-aware evaluation, SFT consistently reduces the expected cost for Qwen2.5-VL-7B by 13.3%, 18.0%, and 21.4% when $w_{\text{over}} = 2, 3, 4$, respectively, and for InternVL3-8B by 4.5%, 8.9%, and 12.1%, demonstrating that the conclusion is stable under different overestimation penalties. For LLaVA1.5-7B, the cost-weighted risk remains unchanged when $w_{\text{over}} = 2$ but decreases when $w_{\text{over}} \geq 3$ (6.3% and 10.7%), indicating that risk-aware gains are also attainable under stricter costs for overestimation.

**Table 10**
Cost-weighted risk (ExpectedCost) computed from confusion matrices with $w_{\text{under}} = 5$ and $w_{\text{over}} \in \{2, 3, 4\}$. Lower is better.

| Backbone | $w_{\text{over}} = 2$ Baseline | SFT | $w_{\text{over}} = 3$ Baseline | SFT | $w_{\text{over}} = 4$ Baseline | SFT |
|---|---|---|---|---|---|---|
| Qwen2.5-VL-7B | 1.109 | 0.961 | 1.333 | 1.093 | 1.558 | 1.225 |
| InternVL3-8B | 1.031 | 0.984 | 1.217 | 1.109 | 1.403 | 1.233 |
| LLaVA1.5-7B | 1.016 | 1.016 | 1.233 | 1.155 | 1.450 | 1.295 |

In terms of generation quality, fine-tuning consistently enhances textual coherence and informativeness, as reflected in the improved BLEU-4 and ROUGE scores across all backbones (Table 8). ROUGE-2, in particular, shows large relative gains, indicating that post-SFT explanations are more detailed and contextually linked rather than short or generic. Qwen2.5-VL-7B achieves the best overall balance across n-gram metrics, InternVL3-8B nearly doubles its ROUGE-2 score, and LLaVA1.5-7B displays the largest relative improvement compared to its weaker baseline. These results confirm that SFT systematically improves both the linguistic quality and content alignment of generated rationales.

Beyond n-gram metrics, an SOTA-based scoring analysis further validates the practical value of these enhancements. For Qwen2.5-VL-7B, Diagnostic Helpfulness (DH) increases from 3.91 to 4.72 (20.7%), Evidence Integration (EI) from 3.01 to 4.20 (39.5%), Contextual Awareness (CA) from 3.89 to 4.68 (20.3%), and Decision Support Value (DSV) from 3.75 to 4.60 (22.7%). InternVL3-8B exhibits comparable trends, with gains of 13.2% in DH, 15.5% in EI, 8.0% in CA, and 14.3% in DSV. LLaVA1.5-7B benefits most in relative terms, showing improvements of 24.9%, 28.4%, 26.4%, and 27.7% across the same dimensions. These results confirm that SFT not only produces more fluent and reference-aligned text but also enhances reasoning quality and real-world decision utility. Collectively, the evidence demonstrates that SFT mean-

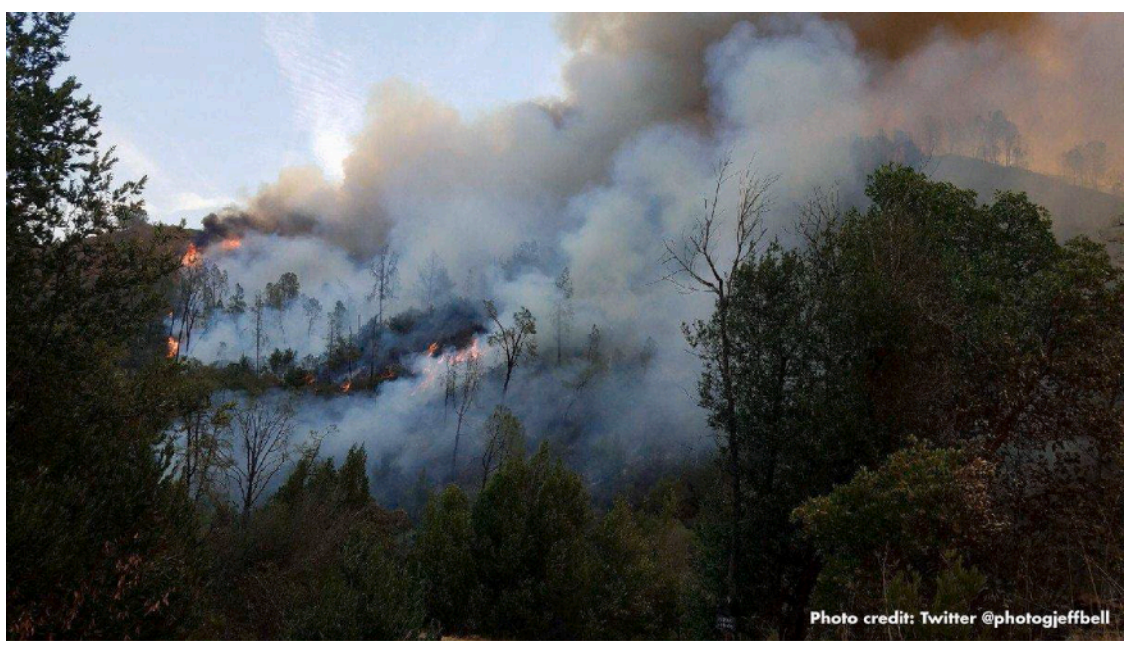

**Tweet_Text:** RT @FAANews: #Pilots: Learn about #FAAa€?s flight restrictions near the #California wildfires. https://t.co/IP5AUbQo6i https://t.co/f9pMEdWLc0

**Label:** severe_damage

**Base:** The image depicts a forest fire with extensive smoke and visible flames. There are no buildings, bridges, or roads in the immediate vicinity that can be assessed for damage. The smoke suggests significant environmental disruption, but the criteria for severe, mild, or little/no damage to infrastructure cannot be applied to this scenario as there is no infrastructure present. The tweet about flight restrictions is unrelated to the image content and does not provide any additional information for damage assessment. Therefore, based on the visual evidence, **the damage level cannot be determined**, and the criteria provided do not apply to this image.

**SFT:** The image depicts a severe wildfire engulfing a forested area, with visible flames and thick smoke indicating extensive environmental damage. While the criteria focus on buildings, bridges, and roads, the intensity of the fire suggests potential severe damage to any nearby infrastructure. The tweet about flight restrictions implies significant air traffic disruption, likely due to hazardous conditions. Given the scale of the fire, it is reasonable to infer that any adjacent buildings, bridges, or roads could be at risk of becoming uninhabitable, unsafe, or impassable, respectively. Therefore, the scene aligns with "**severe_damage**" as the fire's ferocity likely causes widespread destruction beyond the visible area.

**Fig. 7.** Example of baseline refusal on a clear wildfire case using Qwen2.5-VL-7B. The baseline model fails to provide a valid assessment, claiming that no assessable infrastructure is visible, while the SFT-fine-tuned model correctly identifies the scene as severe damage based on flames and dense smoke.

ingfully strengthens both predictive and explanatory capacities across architectures, leading to models that are more accurate, contextually grounded, and operationally useful for crisis response applications.

### *4.2.2. Qualitative observations*

To better understand the nature of improvements achieved through supervised fine-tuning, we focus on the **Qwen2.5-VL-7B** backbone, which serves as the primary implementation in our framework, and analyze representative examples that illustrate key behavioral differences between the baseline and fine-tuned models. Fig. 7 depicts a critical limitation observed in the baseline model when presented with a clear wildfire scene containing visible flames and smoke. The baseline system refused to perform an assessment, stating that "the criteria for severe, mild, or little to no damage to infrastructure cannot be applied to this scenario." This response reflects an overly rigid interpretation of the assessment criteria, overlooking the fact that an active wildfire inherently represents a severe and immediate threat to nearby infrastructure. In contrast, the SFT model correctly classified the scene as severe damage and provided a detailed explanation that explicitly linked the observed fire intensity and smoke patterns to potential structural risks. This richer reasoning demonstrates a more context-aware understanding of visual evidence, aligning closely with the needs of real-world crisis management, where rapid and reliable threat identification is essential.

A second case, shown in Fig. 8, involves a high-risk misclassification scenario in which a museum building is surrounded by floodwater. The baseline model erroneously labeled this case as "little or no damage," justifying its decision solely by noting that the roof structure appeared intact. This shallow reasoning failed to account for flooding effects such as water intrusion, structural compromise, and operational loss. The SFT model, however, correctly identified the situation as severe damage, recognizing that the building was partially submerged and likely unsafe for use. Its explanation emphasized critical factors, including floodwater penetration, potential internal damage, and risks to safety and assets, offering a more holistic interpretation that aligns with crisis response priorities. Such nuanced reasoning is vital in emergency contexts, as misclassifying a heavily flooded site as minimally damaged could delay evacuations, misdirect resources, and result in significant cultural and economic loss.

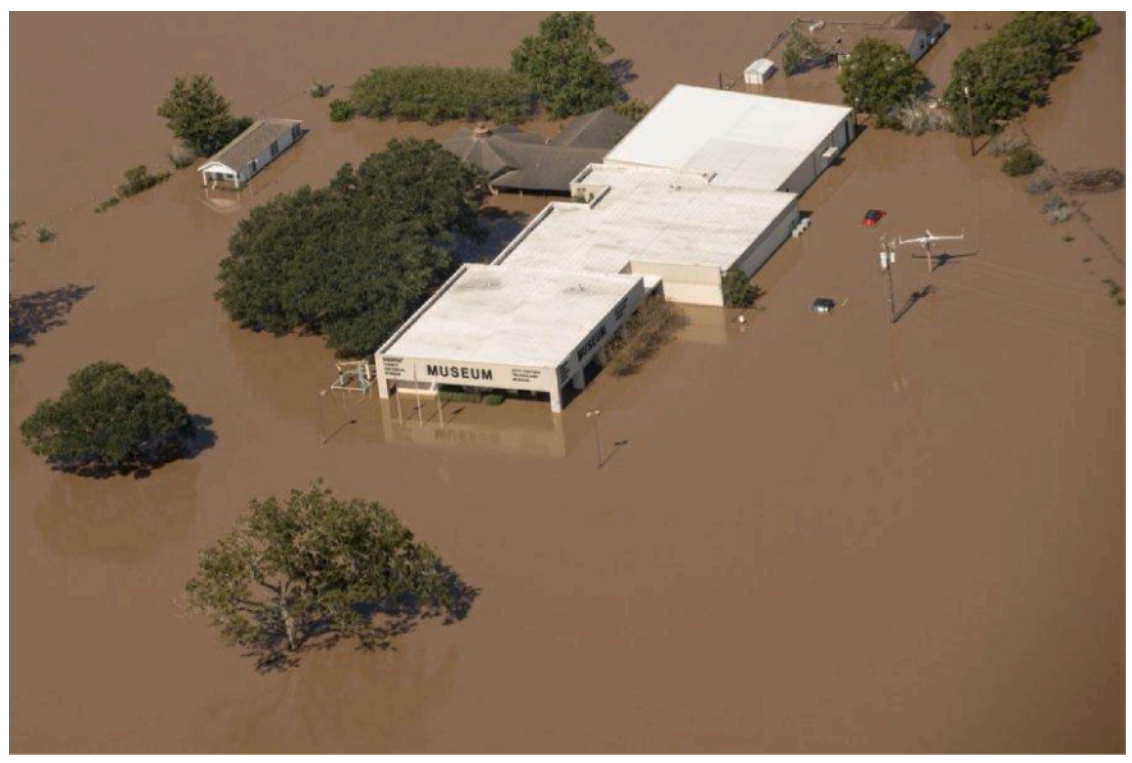

**Tweet_Text:** RT @ideaofthemoment: 12 years ago my parents opened a #museum. A week ago it looked like this á? ¢ 7 #harvey #harveyflood https://t.co/jnT3Ro8Qi5

**Label:** severe_damage

**Base:** The image depicts a severe flood affecting a museum building. The museum's roof appears intact, but the surrounding area is heavily flooded, with water reaching the base of the structure. There are no visible signs of structural damage to the building itself, indicating it may still be usable. However, the surrounding area's condition is indicative of severe flooding, which can lead to significant damage to the museum's exhibits and facilities if not properly managed. The tweet mentions that the museum was opened 12 years ago, suggesting it has been operating for a considerable time, which might be a factor in its current condition. Given the visual evidence, the museum seems to be in a state of **little or no damage** to its structure, but the surrounding area's flooding could pose risks to the museum's contents and operations.

**SFT:** The image depicts severe damage to the museum, aligning with the \"**severe_damage**\" label. The building is partially submerged in floodwaters, indicating significant water intrusion that likely renders it uninhabitable and unsafe. While the structure's roof appears intact, the extensive flooding suggests potential internal damage and structural compromise. The surrounding area is inundated, with no visible roads or bridges, further implying impassability and broader disaster impact. The tweet context reinforces this, as it references a major flood event (Harvey) that caused substantial damage to the museum's operations. Thus, the combination of visual evidence—flooded, potentially unusable facilities—and the historical disaster context confirms severe damage.

**Fig. 8.** Example of high-risk misclassification from the Qwen2.5-VL-7B baseline model. The baseline incorrectly classifies a flooded museum as little or no damage due to the intact roof, overlooking extensive submersion and operational disruption. The SFT-fine-tuned model accurately recognizes severe damage by incorporating evidence of water intrusion, structural compromise, and contextual disaster information.

The qualitative comparison between baseline and fine-tuned outputs highlights two major improvements. First, the SFT model substantially reduces refusal cases by handling a broader range of disaster contexts without deflecting predictions, ensuring more comprehensive coverage across scenarios. Second, it consistently produces richer, evidence-based explanations that integrate visual cues, textual context, and domain-specific reasoning, thereby enhancing both interpretability and decision-support value. Even in instances where both models arrive at the same classification, the SFT version delivers deeper analytical insight by articulating uncertainties, potential secondary impacts, and causal reasoning. Collectively, these enhancements demonstrate that supervised fine-tuning not only improves predictive accuracy but also strengthens the reliability and operational relevance of multimodal reasoning in disaster damage assessment.

## *4.3. Direct preference optimization (DPO) results*

The DPO stage demonstrates that the model effectively leverages human preference annotations to enhance explanation quality and alignment, although the overall improvements over SFT remain modest due to the limited size of the PreferenceSet, which contains 229 training and 29 validation instances. We emphasize that DPO is not intended as a class-imbalance mitigation mechanism; rather, it optimizes preference alignment, and risk-weighted error analysis on the DPO-Testset is inherently high-variance due to its small size (Table 5). As shown in Table 8, Qwen2.5-VL-7B exhibits the most notable progress in minority categories, with the Mild-class F1 increasing from 35.29% to 42.11%, alongside clear improvements in SOTA-based evaluation dimensions such as Diagnostic Helpfulness, Evidence Integration, Contextual Awareness, and Decision Support Value. These enhancements indicate that DPO contributes primarily to improving the interpretability and practical utility of model outputs rather than raw classification accuracy. A similar trend is observed for InternVL3-8B, where Macro-F1 improves from 46.48% to 52.36% and consistent gains are recorded across all four scorer dimensions, while LLaVA1.5-7B also benefits from preference-driven improvements in explanation quality despite minor fluctuations in classification performance. Importantly, the dataset construction process inherently supports this dual-stage approach: high-quality rationales were directly included in the SFT dataset, while lower-quality outputs underwent manual refinement to create human-preference pairs for DPO. This unified data creation pipeline validates the efficiency of deriving both supervised and preference-aligned datasets from a single annotation workflow. Overall, the results emphasize that DPO enhances explanation alignment, interpretability, and decision-support relevance, attributes critical for human trust in crisis assessment systems, even when traditional performance metrics exhibit only moderate gains.

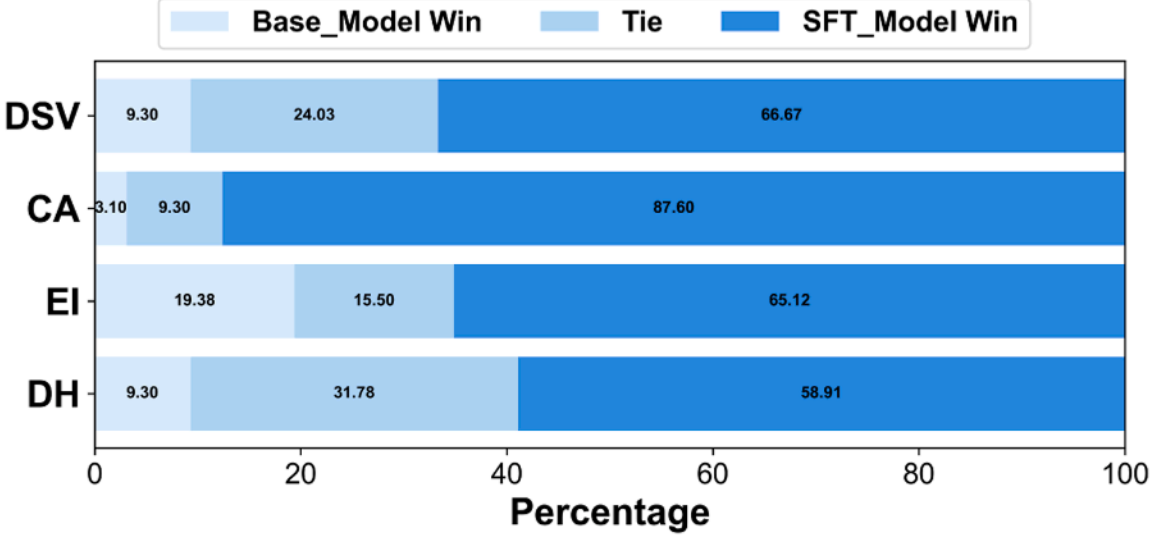

**Fig. 9.** Comparison of human-ranking win rates for Qwen2.5-VL-7B base model and SFT model.

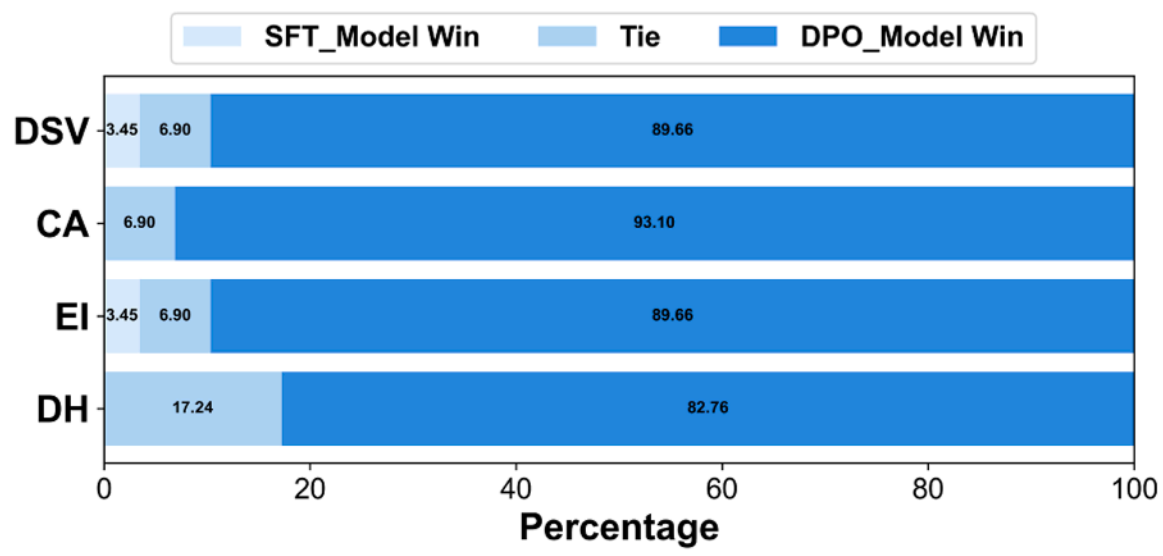

**Fig. 10.** Comparison of human-ranking win rates for Qwen2.5-VL-7B SFT model and DPO model.

**Table 11**
Scorer–human correlation on Qwen2.5-VL-7B (Baseline vs SFT, $n = 129$).

| Dim. | Spearman $\rho$ | $p$ | Agree |
|---|---|---|---|
| DH | 0.398 | $3.06 \times 10^{-6}$ | 0.643 |
| EI | 0.280 | $1.28 \times 10^{-3}$ | 0.667 |
| CA | 0.154 | $8.04 \times 10^{-2}$ | 0.806 |
| DSV | 0.423 | $6.03 \times 10^{-7}$ | 0.713 |
| Overall | 0.386 | $6.26 \times 10^{-6}$ | 0.783 |

**Table 12**
Scorer–human correlation on Qwen2.5-VL-7B (SFT vs DPO, $n = 29$).

| Dim. | Spearman $\rho$ | $p$ | Agree |
|---|---|---|---|
| DH | 0.564 | $1.43 \times 10^{-3}$ | 0.414 |
| EI | 0.310 | $1.02 \times 10^{-1}$ | 0.310 |
| CA | 0.019 | $9.23 \times 10^{-1}$ | 0.276 |
| DSV | -0.093 | $6.31 \times 10^{-1}$ | 0.276 |
| Overall | 0.437 | $1.77 \times 10^{-2}$ | 0.586 |

### *4.4. Human ranking results*

Given the stronger and more stable performance of Qwen2.5-VL-7B observed in earlier experiments, human evaluation was conducted exclusively on this backbone to validate the consistency and practical impact of the proposed framework. The evaluation involved ranking-based comparisons across the four dimensions introduced in the Stage 2 SOTA-based Scorer Assessment, diagnostic helpfulness, evidence integration, contextual awareness, and decision support value, using 129 samples from the SFT test set and 29 from the DPO test set. Annotators were instructed to perform pairwise comparisons instead of assigning absolute scores. As illustrated in Figs. 9 and 10, the results are consistent with automatic and model-based evaluations: the SFT model is consistently preferred over the baseline, demonstrating substantial gains in explanation quality, fluency, and decision relevance, while the DPO-enhanced model is further favored over SFT, reflecting stronger alignment with human expectations. These findings confirm that the two-stage training pipeline progressively enhances both interpretability and practical value, underscoring the significance of integrating human preference signals to produce outputs that are not only accurate but also actionable and trustworthy in real-world crisis management contexts.

### *4.5. Scorer–human consistency*

To assess the robustness of the SOTA-based scorer, we quantify its consistency with human pairwise rankings on the Qwen2.5-VL-7B backbone. For each sample and each dimension (DH/EI/CA/DSV), we compute the scorer margin $\Delta$ as the difference between the two model scores (Baseline vs SFT, or SFT vs DPO). Human ranking outcomes are encoded as $-1/0/+1$, indicating that the first model is preferred, a tie, or the second model is preferred, respectively. We then report Spearman rank correlation [78] between $\Delta$ and the human preference label, as well as the agreement rate (Agree), i.e., the proportion of samples for which the scorer and humans select the same winner/tie. We additionally compute an overall score by averaging the four margins and the four preference labels within each sample.

As shown in Table 11, the scorer exhibits moderate and statistically significant alignment with human judgments for Baseline vs SFT, achieving an overall Spearman correlation of $\rho = 0.386$ ($p = 6.26 \times 10^{-6}$) with 0.783 pairwise agreement. This indicates that the automated ratings are broadly consistent with human preferences when model differences are substantial.

For the more subtle SFT vs DPO comparison (Table 12), the human evaluation is preference-based by design, whereas the scorer produces absolute 1–5 ratings; when the score margins between SFT and DPO are small, this difference in supervision signal can lead to weaker winner-level agreement. Together with the smaller evaluation size ($n = 29$), the dimension-wise consistency becomes noisier than in the Baseline vs SFT setting, even though the overall correlation remains positive ($\rho = 0.437$, $p = 1.77 \times 10^{-2}$) with 0.586 agreement. Overall, these results support using the SOTA-based scorer as an auxiliary evaluation signal, while underscoring the need for complementary human ranking in fine-grained preference comparisons.

#### *4.5.1. Comparative evaluation with published unimodal & multimodal methods*

We compare our approach to representative unimodal (text-only, image-only) and multimodal systems as shown in the Table 13. While unimodal models such as RoBERTa-large and MambaViT can achieve high label accuracy on CrisisMMD splits, and multimodal fusion methods (e.g., CLIP variants) often further increase raw accuracy, these systems are not trained with human-validated rationales and therefore lack the auditable, case-level explanations required for operational decision support. By contrast, our CrisisMMD-R SFT→DPO pipeline explicitly targets explanation quality and preference alignment: it improves BLEU/ROUGE and SOTA-scorer dimensions and obtains favorable human ranking outcomes. This demonstrates that dataset augmentation with human rationales and preference-based alignment produces outputs that are more actionable and trustworthy for crisis response teams, even where classification accuracy gains are incremental.

**Table 13**
Comparison of unimodal, multimodal, and proposed CrisisMMD-R approaches.

| Method | Acc. | M-F1 | BLEU | Explainability |
|---|---|---|---|---|
| RoBERTa-large (Text) | 0.847 | 0.845 | – | No human rationale |
| MambaViT (Vision) | 0.824 | 0.822 | – | No human rationale |
| CLIP Fusion | 0.918 | – | – | Label prediction only |
| Multimodal Aug. | ~0.88 | – | – | No preference alignment |
| General VLMs | – | – | Generated text | Not domain aligned |
| CrisisMMD-R (SFT) | 78.29 | 57.42 | 68.47 | Human-validated rationale |
| CrisisMMD-R (SFT + DPO) | 55.17 | 52.21 | 54.34 | Preference-aligned explanation |

### 4.6. Discussions

In this section, we presented a comprehensive evaluation of our two-stage training approach for disaster damage assessment. The results show that SFT substantially improves both classification accuracy and explanation quality compared with the baseline, particularly for the challenging Mild damage category. Quantitative metrics, SOTA-based scoring, and human ranking consistently indicate that the SFT model not only makes more accurate predictions but also generates richer and more actionable explanations. Qualitative examples further illustrate that SFT reduces refusal cases and high-risk misclassifications, demonstrating more robust and context-aware reasoning. The subsequent DPO stage, while providing more modest gains, helps align model outputs with human preferences and complements the improvements introduced by SFT. Importantly, these patterns are observed consistently across Qwen2.5-VL-7B, InternVL3-8B, and LLaVA1.5-7B, confirming that our two-stage pipeline generalises across architectures. Overall, these findings confirm that combining SFT and DPO in a structured training pipeline can effectively enhance both the reliability and practical utility of multimodal disaster assessment models. The results echo patterns in recent literature. For instance, Ma et al. [79] show that multimodal reasoning pipelines can successfully fuse tweets and images into interpretable assessments that correlate with ground truth seismic signals, reinforcing the value of embedding reasoning capabilities into disaster models. Lei et al. [80] further emphasize that multimodal LLMs improve disaster severity classification when text and visual modalities are jointly leveraged, offering a useful parallel to our gains in classification and explanation. On the preference alignment front, Li et al. [81] demonstrate that DPO can help reconcile modality conflicts and enhance alignment with human expectations in multimodal tasks, providing conceptual support for our observed improvements in explanation quality.

The implications of these findings extend beyond the accuracy of the model and speak directly to the design of next-generation early warning systems. A central requirement in operational disaster management is the ability to embed fine-grained, event-level assessments within a structured and transparent risk governance framework. The INFORM Risk Index provides such a foundation by combining hazard and exposure, vulnerability, and coping capacity into a dynamic measure of humanitarian risk. Our real-time damage estimates and explainable outputs can be mapped to these dimensions to enhance their timeliness and granularity. For instance, severity scores derived from multimodal social media streams can update INFORM's hazard and exposure component. At the same time, behavioral signals such as displacement-related posting patterns may inform vulnerability and coping capacity. The AI-based INFORM extensions described in recent work on early warning systems further support this integration by enabling continuous ingestion of social data, adaptive risk modeling, and automated visualization [82]. Embedding the CrisisMMD-R pipeline within INFORM would therefore enable emergency managers to transition from static annual assessments to continuously updated, spatially resolved risk profiles. This coupling transforms our model from a stand-alone classifier into an operational node of a larger, evidence-based early warning architecture, bridging citizen-sensed micro-level evidence with macro-level risk indices and ensuring that evolving ground realities are rapidly reflected in strategic preparedness and resource allocation.

Equally significant is the behavioral signal embedded in post-disaster user activity. The frequency, timing, and sentiment of user posts often reflect shifting ground realities. For example, surges in requests for medical aid, repeated sharing of shelter locations, or sustained expressions of anxiety can indicate a deterioration of local conditions even when visual evidence is scarce. Incorporating such behavioral markers as auxiliary features could further improve the model's ability to discriminate between mild and severe impacts, especially in situations where direct images of infrastructure damage are unavailable or delayed. The overall design emphasizes decision assistance rather than complete automation. The involvement of human participants in annotating the reasoning output and optimizing the discarded reasoning approaches remains transparent, contestable, and responsive to local context. This human-in-the-loop approach aligns with best practices in high-stakes risk management, where accountability and expert oversight are essential. By generating both a classification and a concise, evidence-based explanation in a single pass, the model delivers actionable information that can be rapidly reviewed and approved within existing command-and-control workflows. These elements together point towards a pathway for real-time operationalization. Integrated into crisis dashboards or early warning platforms, the system could continuously analyze incoming social media streams, highlight areas of probable severe damage, and provide human-readable justifications for immediate verification. Coupled with risk indices such as INFORM and augmented with behavioral signals, this capability would allow decision makers to monitor evolving hazards, prioritize interventions, and allocate resources with greater confidence and speed.

#### 4.6.1. Operational deployment considerations

Although CrisisMMD-R is designed for decision support in time-critical emergency settings, practical deployment requires explicit consideration of runtime and system responsiveness. Training of the final models with LoRA fine-tuning was performed offline and therefore does not affect operational latency. The total training time includes supervised fine-tuning (SFT) followed by Direct Preference Optimization (DPO), since DPO is initialized from the SFT checkpoint. On a single NVIDIA V100-32GB GPU, cumulative training required approximately 2.37 GPU hours for Qwen2.5-VL-7B, 1.29 GPU hours for InternVL3-8B, and 1.86 GPU hours for LLaVA1.5-7B. Because this step is performed only once during model preparation, it does not constrain field usage. At inference time, the model processes a tweet-image pair and produces both a severity label and an explanation. For SFT checkpoints, the average end-to-end response time per instance was 4.43 seconds (Qwen2.5-VL-7B), 3.91 seconds (InternVL3-8B), and 4.82 seconds (LLaVA1.5-7B). For the aligned DPO models, response times increased to 6.97 seconds, 5.38 seconds, and 6.28 seconds, respectively, due to longer reasoning generation. These response times are compatible with emergency monitoring workflows where analysts review incoming streams rather than interact conversationally in real time [27]. Operationally, the system is not intended to trigger actions autonomously. Instead, it functions as a triage assistant, prioritizing items that require human attention. The explanation output allows analysts to verify alignment between model reasoning rapidly and observed evidence, consistent with human-centered AI deployment guidelines [73,83]. This human-review-first design follows reliability engineering principles for high-risk environments, where automated tools augment but do not replace human judgment.

#### *4.6.2. Failure modes and human-in-the-loop safeguards*

Despite alignment training, multimodal reasoning models remain vulnerable to misleading or ambiguous inputs. We observed three recurring failure modes, namely

- **Misleading textual context:** Tweets may contain sarcasm, speculation, or outdated reposted information. Models may generate plausible but incorrect explanations based on linguistic priors rather than visual evidence, a known issue in crisis informatics pipelines [84,85].
- **Ambiguous imagery:** Images depicting debris, smoke, or crowds without clear damage indicators can lead to overestimation of severity. Vision-language models are particularly susceptible to spurious correlations when visual cues are weak [86].
- **Non-disaster or unrelated content:** Posts frequently include unrelated imagery paired with disaster hashtags, producing context hallucinations similar to those observed in large multimodal models [87].

To mitigate these risks, CrisisMMD-R is designed for supervised use with explicit safeguards, such as Explanations displayed alongside predictions for rapid verification, Mandatory review for low-confidence or contradictory reasoning, Preference alignment penalizing unsupported claims, and Human override & audit logging. Such human-in-the-loop workflows are recommended in safety-critical AI systems to maintain accountability and trust calibration [88,89].

#### *4.6.3. Generalizability and cross-domain robustness*

We evaluated multiple architectures (Qwen2.5-VL-7B, InternVL-3-8B, and LLaVA-1.5-7B) to reduce backbone-specific bias. However, generalization in crisis informatics extends beyond model architecture and depends strongly on variations across events and temporal conditions. Disaster communication differs substantially across geography, infrastructure conditions, and reporting culture [90,91]. As a result, models trained on pooled datasets may implicitly learn event-specific correlations rather than transferable reasoning structures. Although the proposed framework encourages evidence-grounded reasoning through human-validated explanations, this study does not explicitly evaluate cross-event or temporal robustness. Prior research suggests that explanation supervision can improve robustness under distribution shift by encouraging models to rely on semantically meaningful features rather than superficial correlations [53]. Nevertheless, crisis data evolves rapidly over time and across contexts [92], and therefore operational deployment would require periodic updating with new human feedback. The design of CrisisMMD-R supports such incremental adaptation, making it compatible with continual learning workflows in operational AI systems [93].

#### *4.6.4. Limitations*

Several limitations must be acknowledged when interpreting the results of this study. The dataset is derived from a single social media platform and a limited set of disaster types, which may restrict transferability to other communication channels, languages, or cultural reporting patterns [90]. In addition, cross-event and temporal generalization were not explicitly evaluated. Future work should examine performance under event holdout settings and temporal shifts to assess robustness in real-world deployments. While inference latency is compatible with analyst-in-the-loop workflows, it may not satisfy the requirements of fully automated alerting systems operating under strict real-time constraints [27]. The system is therefore intended as a decision-support aid rather than an autonomous responder. Human-validated explanations improve interpretability but do not guarantee factual correctness, since annotators may interpret ambiguous imagery differently [94]. Furthermore, even after preference alignment, vision-language models may still produce unsupported causal reasoning when visual evidence is weak [87]. Finally, the study evaluates algorithmic performance rather than full operational deployment. Integration into emergency infrastructures would require governance policies, operator training, monitoring procedures, and long-term trust calibration [73,83].

## 5. Conclusion

This study presents a unified and efficient framework for multimodal disaster damage assessment that advances both methodological rigor and practical applicability in reliability-oriented contexts. The central contribution lies in the design of a novel human-in-the-loop data construction pipeline, which simultaneously produces two complementary datasets, ReasoningSet for SFT and PreferenceSet for DPO, within a single annotation workflow. This integrated approach significantly reduces annotation costs and time while providing distinct yet synergistic supervision signals for a two-stage model adaptation process. By consolidating data collection and aligning human refinement with model training, the framework enables dual-purpose supervision without redundancy, an essential feature for emergency management systems operating under resource and time constraints.

Methodologically, SFT enables the model to jointly learn disaster classification and rationale generation, resulting in notable gains in accuracy, robustness, and interpretability. The approach mitigates high-risk misclassifications, reduces model refusals, and substantially improves recognition of the underrepresented Mild damage category. Building on this foundation, DPO introduces preference-based alignment that integrates human judgments to refine the contextual and ethical quality of generated rationales. This alignment not only strengthens interpretability but also provides an auditable reasoning trail, a crucial component for accountability and traceability in emergency decision-making. Complementing these methodological contributions, a comprehensive three-tiered evaluation framework encompassing automatic metrics, state-of-the-art model-based scoring, and human ranking experiments demonstrates the reliability and operational utility of the proposed system. From an applied perspective, this research contributes to the broader goal of developing trustworthy AI systems for crisis response. The lightweight Gradio-based prototype illustrated in Fig. 5 exemplifies how the resulting models can be integrated into human-in-the-loop decision-support environments, providing real-time explainable outputs that facilitate coordination, validation, and after-action review. Importantly, the observed improvements generalize across multiple backbones, Qwen2.5-VL-7B, InternVL3-8B, and LLaVA1.5-7B, confirming the flexibility and scalability of the pipeline.

While the dataset curation process includes expert verification and correction, we did not perform a quantitative severity-weighted error risk analysis. We didn’t explicitly measure the frequency of high-impact misclassification types (e.g., severe underestimation) nor compute a formal risk-weighted penalty reflecting operational consequences. Instead, the current work focuses on improving annotation validity through professional adjudication and preference-pair construction. Future work will extend the evaluation by incorporating safety-oriented metrics that quantify the impact of different error categories. This includes measuring the distribution of severity-level disagreements and weighting them based on operational risk, as commonly recommended in decision-support system evaluation. Such analysis would enable a more direct assessment of how model behavior affects emergency response prioritization beyond standard accuracy and explanation quality metrics. The CrisisMMD-R dataset, while carefully curated, is limited in scale and diversity, which may constrain generalization across varied disaster types or linguistic contexts. Computational resources also restricted experimentation to mid-sized models (7B/8B), leaving the performance of larger architectures (e.g., 70B) for future exploration. Moreover, while DPO yielded meaningful improvements in explanatory alignment, further work is needed to compare alternative preference optimization strategies and assess robustness under dynamic, real-world conditions. Looking ahead, future research will focus on extending this framework toward a fully integrated, reliable emergency management decision-support system. Key directions include scaling preference datasets via crowd-assisted annotation, integrating real-time social and remote-sensing data streams, and linking model outputs to operational dashboards for situational awareness and resource allocation. Integration with early-warning systems and incident-management platforms could enable continuous feedback loops between automated assessments and practitioner interventions. Additionally, embedding uncertainty estimation, bias monitoring, and provenance tracking will be essential for ensuring transparency and trustworthiness in high-stakes scenarios. Through these developments, the proposed framework can evolve into a cornerstone for next-generation AI systems that provide reliable, explainable, and ethically grounded decision support for emergency management and disaster resilience.

## CRediT authorship contribution statement

**Yuanjun Zhang:** Writing – review & editing, Writing – original draft, Visualization, Methodology, Data curation, Conceptualization; **Fuzel Ahamed Shaik:** Writing – review & editing, Visualization, Project administration, Methodology, Conceptualization; **Suvojit Acharjee:** Software; **Fahad Khalid:** Writing – review & editing; **Mourad Oussalah:** Writing – review & editing, Supervision.

## Data availability

To minimize risks of misuse, access to the derived text datasets will be provided under a research-only license. Requests can be made via institutional email, and applicants must agree not to use the data for re-identification of individuals or for any commercial purpose. This work is built on CrisisMMD and does not directly release raw content from Platform X. We will not redistribute the original CrisisMMD media or any Platform X data. Instead, we will provide only an incremental layer aligned to CrisisMMD samples. Researchers must obtain CrisisMMD through its original distribution channel and comply with its licensing terms and any applicable Platform X policies when hydrating or accessing underlying posts. This incremental layer includes ReasoningSet, PreferenceSet and the train/dev/test split indices used in our experiments. Data will be made available on request.

## Declaration of competing interest

The authors declare that they have no known competing financial interests or personal relationships that could have appeared to influence the work reported in this paper.

## Appendix A. Evaluation Stability under an Alternative Judge Model

Table A.14 evaluates the stability of our explanation scoring by swapping the judge model from Qwen-VL-Max to GPT-4o-mini while keeping the evaluation protocol unchanged (same test sets, the same four dimensions DH/EI/CA/DSV, and the same judging prompt/template). Pink rows correspond to the SFT-Testset and blue rows to the DPO-Testset; within each backbone, the best SFT scores are in **bold** and the best DPO scores are underlined (applied independently for each judge block). Across all three backbones, both judges agree on the main ranking on the SFT-Testset: *SFT consistently outperforms the Baseline on all four dimensions*. On the DPO-Testset, both judges also agree that *SFT + DPO is the best or tied-best variant for most dimensions*, suggesting the preference alignment does not introduce brittle, judge-specific gains: for Qwen2.5-VL-7B, SFT + DPO improves over SFT under both judges; for InternVL3-8B, SFT + DPO again improves consistently. The only minor deviation is LLaVA1.5-7B on EI under Qwen-VL-Max where SFT is slightly higher (SFT vs. SFT + DPO: 4.41 vs. 4.34), while GPT-4o-mini favors SFT + DPO (SFT vs. SFT + DPO: 4.17 vs. 4.34); importantly, this does not flip the overall conclusion. Since absolute score scales can differ between judges, we focus on the stability of relative comparisons within each backbone and test set; overall, the qualitative findings remain consistent under the alternative judge model.

**Table A.14**
Evaluation stability with an alternative judge model (Qwen-VL-Max vs. GPT-4o-mini). Within each backbone, the best SFT scores are in **bold**, while the best DPO scores are underlined.

| Backbone | Method | Dataset | Qwen-VL-Max | | | | GPT-4o-mini | | | |
|---|---|---|---|---|---|---|---|---|---|---|
| | | | DH | EI | CA | DSV | DH | EI | CA | DSV |
| Qwen2.5-VL 7B-Instruct | Baseline | SFT-Testset | 3.91 | 3.01 | 3.89 | 3.75 | 4.33 | 4.02 | 4.40 | 4.05 |
| | SFT | | **4.72** | **4.20** | **4.68** | **4.60** | **4.34** | **4.23** | **4.47** | **4.21** |
| | SFT | DPO-Testset | 4.34 | 4.31 | 4.38 | 4.28 | 4.14 | 4.03 | 4.31 | 3.86 |
| | SFT + DPO | | 4.59 | 4.52 | 4.38 | 4.41 | 4.21 | 4.17 | 4.38 | 4.07 |
| InternVL 3-8B | Baseline | SFT-Testset | 4.10 | 3.88 | 4.26 | 4.00 | 4.19 | 3.97 | 4.44 | 3.98 |
| | SFT | | **4.64** | **4.48** | **4.60** | **4.57** | **4.26** | **4.21** | **4.52** | **4.21** |
| | SFT | DPO-Testset | 4.38 | 4.34 | 4.21 | 3.97 | 4.21 | 4.03 | 4.28 | 3.90 |
| | SFT + DPO | | 4.48 | 4.45 | 4.24 | 4.00 | 4.34 | 4.41 | 4.41 | 4.14 |
| LLaVA 1.5-7B | Baseline | SFT-Testset | 3.78 | 3.48 | 3.53 | 3.54 | 3.70 | 3.47 | 3.99 | 3.43 |
| | SFT | | **4.72** | **4.47** | **4.46** | **4.52** | **4.29** | **4.25** | **4.49** | **4.13** |
| | SFT | DPO-Testset | 4.45 | 4.41 | 4.14 | 4.14 | 4.28 | 4.17 | 4.41 | 4.14 |
| | SFT + DPO | | 4.48 | 4.34 | 4.17 | 4.17 | 4.28 | 4.34 | 4.41 | 4.24 |

## Appendix B. Reproducibility Details

Training and inference are both conducted using the LLaMA-Factory [22] framework ($\geq$ 0.9.4); this paper does not provide a basic tutorial on LLaMA-Factory [22]. Unless otherwise specified, any parameters not explicitly stated use the framework's default values. Trained on NVIDIA V100 32GB. This appendix summarizes all information needed to reproduce our pipeline: the prompts (Figs. B.11 and B.12), the ShareGPT data format for SFT/DPO (Listings 3 and 4), and the full training configurations for SFT/DPO in LLaMA-Factory (Listings 5 and 6).

**Prompt for Visual-Tweet Reasoning to Generate a Draft**

CoT_PROMPT = """You are a disaster damage assessment expert.

**Criteria:**
Severe Damage:
- Buildings: Uninhabitable/unusable, large areas of roof missing, structurally unsafe
- Bridges: Clearly unsafe, partially collapsed, impassable
- Roads: Impassable, with boulder accumulation, large potholes, or severe road subsidence

Mild Damage:
- Buildings: Partially damaged but still usable, no more than 50% of roof/facilities damaged
- Bridges: Still usable but some areas need repair
- Roads: Still passable but some sections need to be closed

Little or No Damage:
- Infrastructure basically intact, only normal aging and wear

**Task:** Analyze the image and tweet to assess damage level according to the above criteria. Tweets should only be used as auxiliary judgment; the core focus should be on image assessment. Emphasis should be placed on the reasoning process.

**Tweet: {tweet_text}**
**Known label: {damage_level}**

**Output** only the final integrated reasoning (combine visual evidence and tweet context into a coherent explanation of why this is **{damage_level}**) in 100-150 words:"""

**Fig. B.11.** Prompt for visual-tweet reasoning to generate a draft.

**You are a disaster damage assessment expert.**

*Criteria*
**Severe Damage** - Buildings: Uninhabitable/unusable, large areas of roof missing, structurally unsafe - Bridges: Clearly unsafe, partially collapsed, impassable - Roads: Impassable, with boulder accumulation, large potholes, or severe road subsidence
**Mild Damage** - Buildings: Partially damaged but still usable, no more than 50% of roof/facilities damaged - Bridges: Still usable but some areas need repair - Roads: Still passable but some sections need to be closed
**Little or No Damage** - Infrastructure basically intact, only normal aging and wear

*Task*
Analyze the image and tweet to assess damage level according to the above criteria. Tweets should only be used as auxiliary judgment; the core focus should be on image assessment. Emphasis should be placed on the reasoning process. Provide integrated reasoning that combines visual evidence and tweet context into a coherent explanation in 100-150 words.

**Fig. B.12.** Fixed system prompt used in all experiments.

```
# Construct ShareGPT format in SFT stage
sample = {
    "conversations": [
        {
            "from": "human",
            "value": f"<image>{system_prompt}
                Tweet: '{row['tweet_text']}'"
        },
        {
            "from": "gpt",
            "value": row['final_response']
        }
    ],
    "images": [image_path]
}
```

**Listing 3.** SFT ShareGPT

```
# Construct ShareGPT format in DPO stage
sample = {
    "conversations": [
        {
            "from": "human",
            "value": f"<image>{system_prompt}
                Tweet: '{row['tweet_text']}'"
        },
    ],
    "chosen": {
        "from": "gpt",
        "value": row['modified_response']
    },
    "rejected": {
        "from": "gpt",
        "value": row['final_response']
    },
    "images": [image_path]
}
```

**Listing 4.** DPO ShareGPT

```yaml
### model
model_name_or_path: your_path
adapter_name_or_path: null
### method
stage: sft
do_train: true
finetuning_type: lora
lora_target: all
lora_rank: 16
lora_alpha: 32
lora_dropout: 0.1
### freeze
freeze_vision_tower: true
freeze_multi_modal_projector: true
### dataset
dataset: your_dataset_name
eval_dataset: your_dataset_name
template: your_model_template_name
cutoff_len: 2048
overwrite_cache: true
preprocessing_num_workers: 8
### output
output_dir: your_path
logging_steps: 13
save_steps: 43
plot_loss: true
overwrite_output_dir: true
report_to: tensorboard
logging_dir: your_path
### train
per_device_train_batch_size: 1
gradient_accumulation_steps: 8
learning_rate: 7.0e-5
num_train_epochs: 3
lr_scheduler_type: cosine
warmup_ratio: 0.1
bf16: false
fp16: true
ddp_timeout: 180000000
optim: adamw_torch
### eval
do_eval: true
per_device_eval_batch_size: 1
eval_strategy: steps
eval_steps: 43
metric_for_best_model: eval_loss
greater_is_better: false
load_best_model_at_end: true
save_total_limit: 5
```

**Listing 5.** SFT yaml

```yaml
### model
model_name_or_path: your_path
adapter_name_or_path: your_path
### method
stage: dpo
do_train: true
finetuning_type: lora
lora_target: all
lora_rank: 16
lora_alpha: 32
lora_dropout: 0.1
### freeze
freeze_vision_tower: true
freeze_multi_modal_projector: true
### dataset
dataset: your_dataset_name
eval_dataset: your_dataset_name
template: your_model_template_name
cutoff_len: 2048
overwrite_cache: true
preprocessing_num_workers: 8
### output
output_dir: your_path
logging_steps: 5
save_steps: 10
plot_loss: true
overwrite_output_dir: true
report_to: tensorboard
logging_dir: your_path
### train
per_device_train_batch_size: 1
gradient_accumulation_steps: 8
learning_rate: 5.0e-6
num_train_epochs: 4
lr_scheduler_type: cosine
warmup_ratio: 0.1
bf16: false
fp16: true
ddp_timeout: 180000000
optim: adamw_torch
### eval
do_eval: true
per_device_eval_batch_size: 1
eval_strategy: steps
eval_steps: 10
metric_for_best_model: eval_loss
greater_is_better: false
load_best_model_at_end: true
save_total_limit: 3
```

**Listing 6.** DPO yaml